\documentclass[letterpaper, 10 pt, journal, twoside]{ieeetran}
\IEEEoverridecommandlockouts

\usepackage{amsmath,amssymb, graphicx, epstopdf,cite,enumerate,booktabs, setspace, float,stfloats,bm, multirow, lscape, caption, subcaption, graphbox, soul, color, xcolor, hyperref}

\hypersetup{hidelinks}

\usepackage[normalem]{ulem}
\usepackage[linesnumbered,ruled]{algorithm2e}
\usepackage[bottom=2.4cm,top=2.01cm,left=1.69cm,right=1.69cm]{geometry}

\title{RDA: An Accelerated Collision Free\\Motion Planner for Autonomous Navigation in\\Cluttered Environments}

\author{Ruihua Han$^{1,2}$, Shuai Wang$^{3}$, Shuaijun Wang$^{5, 1}$, Zeqing Zhang$^{2}$, Qianru Zhang$^{2}$,\\Yonina C. Eldar$^4$, Qi Hao$^{1,\dagger}$, Jia Pan$^{2,\dagger}$

\thanks{$^{1, 2}$ R. Han is with the Department of Computer Science and Engineering, Southern University of Science and Technology, Shenzhen, Guangdong, China, and also with the Department of Computer Science, The University of Hong Kong, Hong Kong 
  {\tt\footnotesize hanrh@connect.hku.hk}}
\thanks{$^{3}$ S. Wang is with Shenzhen Institute of Advanced Technology, Chinese Academy of Sciences, Shenzhen, Guangdong, China}
\thanks{$^{5}$ S. Wang is with the Department of Computer Science and Engineering, Harbin Institute of Technology, Shenzhen, Guangdong, China}
\thanks{$^{2}$ Z. Zhang, Q. Zhang, and J. Pan are with the Department of Computer Science, The University of Hong Kong, Hong Kong 
  {\tt\footnotesize jpan@cs.hku.hk}}
\thanks{$^{4}$ Y. C. Eldar is with the Weizmann Institute of Science, Rehovot, Israel}
\thanks{$^{1}$ Q. Hao is with the Department of Computer Science and Engineering, the Shenzhen Key Laboratory of Robotics and Computer Vision, and the Sifakis Research Institute for Trustworthy Autonomous Systems, Southern University of Science and Technology, Shenzhen, Guangdong, China
  {\tt\footnotesize hao.q@sustech.edu.cn}}
\thanks{$^\dagger$ denotes the corresponding authors}
} 

\begin{document}
\maketitle

\begin{abstract}
Autonomous motion planning is challenging in multi-obstacle environments due to nonconvex collision avoidance constraints. Directly applying numerical solvers to these nonconvex formulations fails to exploit the constraint structures, resulting in excessive computation time. In this paper, we present an accelerated collision-free motion planner, namely regularized dual alternating direction method of multipliers (RDADMM or RDA for short), for the model predictive control (MPC) based motion planning problem. The proposed RDA addresses nonconvex motion planning via solving a smooth biconvex reformulation via duality and allows the collision avoidance constraints to be computed in parallel for each obstacle to reduce computation time significantly. We validate the performance of the RDA planner through path-tracking experiments with car-like robots in both simulation and real-world settings. Experimental results show that the proposed method generates smooth collision-free trajectories with less computation time compared with other benchmarks and performs robustly in cluttered environments. The source code is available at \href{https://github.com/hanruihua/RDA_planner}{\url{https://github.com/hanruihua/RDA_planner}}. 

\end{abstract}

\begin{IEEEkeywords}
Collision avoidance, constrained motion planning, optimization and optimal control
\end{IEEEkeywords}

\section{Introduction}\label{section1}

\IEEEPARstart{M}{otion} planning has drawn rising interest in many practical applications, such as self-driving cars and unmanned logistic vehicles~\cite{zhang2021efficient, sonneberg2019autonomous}. Its main idea is to compute a trajectory, defined as a sequence of control commands or states, from the current state to the desired state. Motion planners can be classified into four categories~\cite{gonzalez2015review}: graph search based, sampling based, interpolation based, and optimization based. Graph search based techniques, e.g., state lattice and A star~\cite{pivtoraiko2009differentially, montemerlo2008junior}, treat the configuration space as a graph of vertices and edges and search the minimum cost path. Sampling based techniques, e.g., the rapidly-exploring random tree (RRT)~\cite{radaelli2014motion}, probe the configuration space with a sampling scheme and have been widely used in real-world applications. Interpolation based techniques construct a new group of data among known reference points through interpolation models, resulting in a feasible, smooth trajectory in structured environments~\cite{funke2012up}. 
Compared with other methods, optimization based techniques are advantageous in generating optimized and robust trajectories in complex scenarios through minimizing cost functions under physical constraints, such as dynamics, kinematics, and collision avoidance. 
For instance, the model predictive control (MPC) approach adopts the vehicle dynamics model to predict future states and produce feasible trajectories dynamically within a set of constraints~\cite{ammour2021collision}. 

Despite its promising features, there are two challenges hindering the development of optimization based techniques in real-world applications.
First, the motion planning optimization problem is usually nonconvex due to collision avoidance constraints. Numerous approaches have been proposed to resolve this issue, such as linearizing the constraints by Taylor expansion~\cite{rey2018fully, wang2018parallel}, relaxing the constraints to formulate an unconstrained problem~\cite{wang2021geometrically}, and reformulating the constraints via strong duality~\cite{zhang2020optimization, xia2022trajectory}. However, most of these approaches consider the point-mass object model, or involve large approximation errors, or fail to exploit the problem structure. 
The effectiveness of optimization methods under the non-point-mass model requires further investigation (i.e., accounting for the object shape).
Second, the computation cost grows significantly with the number of obstacles.  
While several parallel algorithms~\cite{cheng2021admm} have been developed to split multi-agent planning problems into smaller subproblems for each agent, they cannot be directly extended to the parallelization of obstacles. 
On the other hand, the primal dual alternating optimization (PDAO) in{~\cite{firoozi2020distributed}} is able to update different collision avoidance constraints in parallel.
However, PDAO may diverge due to the nonlinear coupling among primal and dual variables in collision avoidance constraints. 
Additionally, PDAO ignores the different importance of upcoming and future states due to the same safety distance for all time slots.
Hence, developing a parallel and adaptive computing framework to effectively reduce the computation cost becomes imperative.

This paper proposes an accelerated collision-free motion planner that realizes accurate, real-time, and robust navigation under real-world obstacle shapes, computation constraints, and environmental uncertainties. 
The main contributions are summarized as follows.
\begin{itemize}
\item[1)] We formulate an optimization based navigation problem with the non-point-mass obstacle model. The nonconvex constraints are reformulated into bi-convex counterparts by joint linearization and strong duality. The safety distance in the collision avoidance constraints is automatically tuned via $l_1$ regularization.

\item[2)] We develop a parallel local planner, termed regularized dual alternating direction method of multipliers (RDA), to solve the bi-convex problem in parallel while bringing robustness to the algorithm convergence.

\item[3)] We implement the RDA algorithm in a high-fidelity Gazebo simulator and an Ackermann autonomous vehicle, respectively, and compare its performance with several benchmark schemes in terms of computation cost and success rate. 
\end{itemize}

The remainder of this paper is organized as follows. Section~{\ref{section2}} reviews the related work. Section {\ref{section3}} introduces the distance based collision avoidance constraint and optimization problem statement. Then, Section~{\ref{section4}} describes the proposed RDA collision-free motion planner. Section~{\ref{section5}} describes the simulation and real-world experiments. Section~{\ref{section6}} presents the conclusion.

\section{Related Work}\label{section2} 

\begin{figure*}[t]
  \centering
    \includegraphics[width=0.7\textwidth]{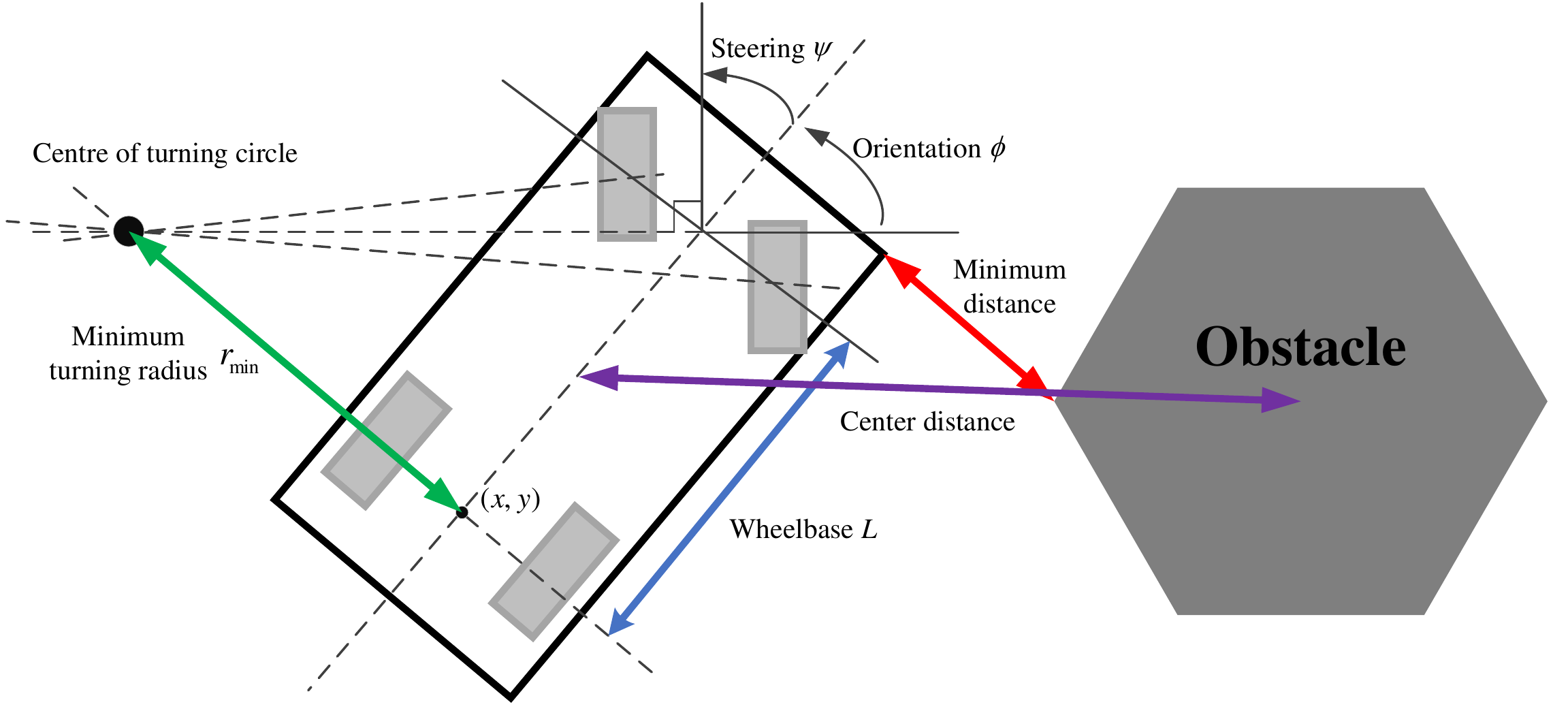}
    \caption{Illustration of the Ackerman kinetic model and the distance between the robot and the obstacle.}
  \label{Ackermann}
\end{figure*}

Optimization based collision-free motion planners aim to minimize or maximize a utility function w.r.t. a set of constrained state and control variables. 
For instance, in \cite{funke2016collision}, an MPC scheme is proposed to accomplish the path tracking task by using the varied length time in the prediction horizon, where the collision avoidance constraint is modeled by the environmental envelope~\cite{erlien2013safe}. 
In fact, the time elastic bands (TEB) approach, which is the most popular method in robotics navigation, is also optimization based, which realizes motion planning by solving a multi-objective optimization problem~\cite{rosmann2017kinodynamic}. 
To guarantee collision avoidance, a formulation of the constraint based on the minimum distance between robots and obstacles has been proposed, which, however, is nonconvex~\cite{rodriguez2014trajectory, zhang2022generalized}. 
Duality based approaches have been studied recently to handle nonconvexity, where the distance functions are reformulated to be smooth, differentiable counterparts via strong duality. The ``least-intrusive'' trajectory method is proposed in~\cite{zhang2020optimization} for non-point-mass objects. However, its final formulation involves bi-linear terms and needs to be solved via nonconvex numerical solvers. 
The work~\cite{xia2022trajectory} linearizes the bi-linear terms through duality and develops a sequential convex programming method. 
Nevertheless, only point-mass objects are considered in \cite{xia2022trajectory}, which is inappropriate for practical polygonal robots such as self-driving cars. 
Furthermore, the complexity of sequential convex programming is exceedingly high when the number of obstacles is large.

There are many ways to reduce the computation cost for solving the large-scale motion planning problem.
Specifically, we could consider only the adjacent agents and obstacles within a pre-defined range of interest such that
the number of agents and obstacles is significantly reduced.
However, the range of interest increases with the velocity of the vehicle, and collisions may occur if some out-of-range obstacles suddenly appear or accelerate. 
Distributed optimization has been leveraged to solve the problem in parallel, e.g., collision avoidance in multi-robot systems~\cite{firoozi2020distributed}. 
However, the algorithm in{~\cite{firoozi2020distributed}} is not compatible with the real-time robot operation system (ROS). Thus, the performance of parallel computing under practical sensor and actuator uncertainties requires further investigation.

As a promising distributed optimization framework, the alternating direction method of multipliers (ADMM)~\cite{boyd2011distributed} has been adopted in multi-agent systems~\cite{rey2018fully, wang2018parallel, cheng2021admm}, which decompose a large centralized problem into several subproblems solved for each agent.
However, for the cluttered environment with multiple obstacles, decomposing the problem w.r.t each obstacle via ADMM has not yet been investigated.
The method adopted in multi-agent systems cannot be applied directly to multi-obstacle systems, since different agents have their own motion dynamics. The collision avoidance constraints are nonlinearly coupled with each other by a single motion dynamic constraint. 
Here, we tackle this coupling issue via joint linearization and duality and then leverage regularization and ADMM to enhance the robustness of the algorithm.


\section{Problem Statement}\label{section3} 

\subsection{Non-Point-Mass Collision Avoidance Constraint}

The point-mass object is not a good model for some specific robots. For instance, for a robot with Ackermann kinematic model, as shown in Fig. \ref{Ackermann}. It is vital to take the rectangle shape into account when navigating in the corridor. Thus, a non-point-mass (i.e. arbitrary convex shape) representation is necessary.

\emph{1) Obstacle Model:}
The environment consists of $M$ non-point-mass obstacles. The $m$th obstacle ($m=1,\cdots,M$) is modeled as a compact convex set $\mathbb{O}_m$ that is represented by the conic inequality~\cite{boyd2004convex}: 
\begin{equation}
     {\mathbb{O}_m} = \{ \mathbf{o} \in {\mathbb{R}^{o_m}}|\mathbf{D}_m\mathbf{o}{ \preceq _{\mathcal{O}_m}} \mathbf{b}_m\}, 
    \label{om}
\end{equation}
where $\mathbf{D}_m \in {\mathbb{R}^{l_m \times o_m}}$, $\mathbf{b}_m \in {\mathbb{R}^{l_m}}$, and $\mathcal{O}_m$ is a proper cone. The partial ordering w.r.t. a cone $\mathcal{K}$ is defined by $\mathbf{x}{ \preceq _\mathcal{K}}\mathbf{y} \Leftrightarrow \mathbf{y} - \mathbf{x} \in \mathcal{K}$. 
Each geometric object can be represented with a proper cone through this formula.
For example, if $\mathcal{O}_m{\rm{ = }}{\mathbb{R}_ + }$, the obstacle is a polyhedron; if $\mathcal{O}_m = \{ (\mathbf{x},t) \in {\mathbb{R}^{l_m + 1}}|{\left\| \mathbf{x} \right\|_2} \le t\}$, the obstacle is an ellipsoid. For static obstacles, the set $\mathbb{O}$ is a constant; for dynamic obstacles, the set $\mathbb{O}_m$ can be denoted as $\mathbb{O}_{m}^t$, which changes over time.

\emph{2) Robot Model:}
The state of each robot at time $t$ is represented by its center point and denoted as $\mathbf{s}_t\in \mathbb{R}^{n_r}$. The value of $n_r$ depends on the working space dimension. 
When the robot moves in a plane, we have $n_r=3$ and $\mathbf{s}_t=(x_t,y_t,\theta_{t})$, 
where $(x_t,y_t)$ represents the position and $\theta_{t}$ the orientation.
Given a certain state, the robot can be modeled as a compact set $\mathbb{Z}_t$ changing with state variable $\mathbf{s}_t$:
\begin{align}
  \mathbb{Z}_t(\mathbf{s}_t) &= \mathbf{R}(\mathbf{s}_t)\mathbf{z} + \mathbf{p}(\mathbf{s}_t),~\forall\mathbf{z}\in\mathbb{C}, \nonumber
  \\
  \mathbb{C}&= \{ \mathbf{z} \in {R^{n_r}}| \mathbf{G}\mathbf{z}{ \preceq _{\mathcal{K}_r}}\mathbf{h}\}, 
\end{align}
where $\mathbf{R}(\mathbf{s}_t) \in {\mathbb{R}^{{n_r} \times {n_r}}}$ is the rotation matrix representing the orientation of the robot and $\mathbf{p}(\mathbf{s}_t) \in {\mathbb{R}^{{n_r}}}$ is the transition matrix representing the position of the robot. 
The vector $\mathbf{z} \in {\mathbb{R}^{{n_r}}}$ is the initial pose vector, and $\mathbb{C}$ is a convex set representing the robot shape at the initial position, with $\mathbf{G} \in {R^{h \times n_r}}$ and $\mathbf{h} \in {\mathbb{R}^h}$. 
We assume $\mathbf{R}(\mathbf{s}_t)$ and $\mathbf{p}(\mathbf{s}_t)$ are linear functions of $\mathbf{s}$. Any nonlinear function is represented by a linear approximation. For instance, the rotation matrix can be linearized by its Taylor expansion \cite{barfoot2011pose}. 

\emph{3) Collision Avoidance:}
The minimum distance between a robot and an obstacle ${\bf{dist}}(\mathbb{Z}_t(\mathbf{s}_t), \mathbb{O})$ satisfies:
\begin{align}
  {\bf{dist}}(\mathbb{Z}_t(\mathbf{s}_t), \mathbb{O}) = \min \left\{ {\left. {{{\left\| \mathbf{e} \right\|}_2}} \right|(\mathbb{Z}_t(\mathbf{s}_t) + \mathbf{e}) \cap \mathbb{O} \ne \emptyset } \right\}, 
  \label{dist}
\end{align}
where $\emptyset$ denotes the empty set. Substituting the expressions of $\mathbb{Z}_t(\mathbf{s}_t)$ and $\mathbb{O}$ into {\eqref{dist}}, we can find ${\bf{dist}}(\mathbb{Z}_t(\mathbf{s}_t), \mathbb{O})$ by solving the following convex problem:
\begin{equation}
\begin{aligned}
    \mathop {{\rm{min}}}\limits_{\mathbf{z},\mathbf{o}}~&{\left\| {\mathbf{R}(\mathbf{s}_t)\mathbf{z} + \mathbf{p}(\mathbf{s}_t) - \mathbf{o}} \right\|_2}
    \\
    \text { s.t. }~&\mathbf{D}\mathbf{o}{ \preceq _{{\mathcal{K}_o}}}\mathbf{b},~ \mathbf{G}\mathbf{z}{ \preceq _{{\mathcal{K}_r}}}\mathbf{h}
    \label{opt}
\end{aligned}
\end{equation} 
with variable $\mathbf{z}$ and $\mathbf{o}$. Thus, to guarantee the collision avoidance, we have the following constraint:
\begin{equation}
{\bf{dist}}(\mathbb{Z}_t(\mathbf{s}_t),\mathbb{O}) \ge {d_{\mathrm{safe}}},
\label{cac}
\end{equation}
where $d_{\mathrm{safe}}$ is a positive real number representing the safe distance. Constraint ({\ref{cac}}) is nonconvex as the convex distance function is at the left hand side of operator $\geq$.

\subsection{State Evolution}

The robot state evolution can be described as follows:
\begin{equation}
    {\mathbf{s}_{t + 1}} = {\mathbf{s}_t} + f({\mathbf{s}_t},{\mathbf{u}_t})\Delta t, 
    \label{dyna}
\end{equation}
where the initial state ${\mathbf{s}_{0}}$ is the current state, $\mathbf{u}_t\in {\mathbb{R}^{{n_u}}}$ is the control vector, and $\Delta t$ is the time length of each slot. When the robot moves in a plane, we have $n_u=2$ and $\mathbf{u}_t=(v_t, \psi_t)$, 
where $(v_t,\psi_t)$ represents the linear and angular velocities.
\footnote{
The linear and angular velocities ($v_t$, $\psi_t$) may also be added to the state space model, where we have $n_r=5$ and accelerations become the control vector.
Note that this would lead to an equivalent optimization problem, and the proposed RDA is still applicable. 
}

\begin{figure}[t]
  \centering
      \includegraphics[width=0.45\textwidth]{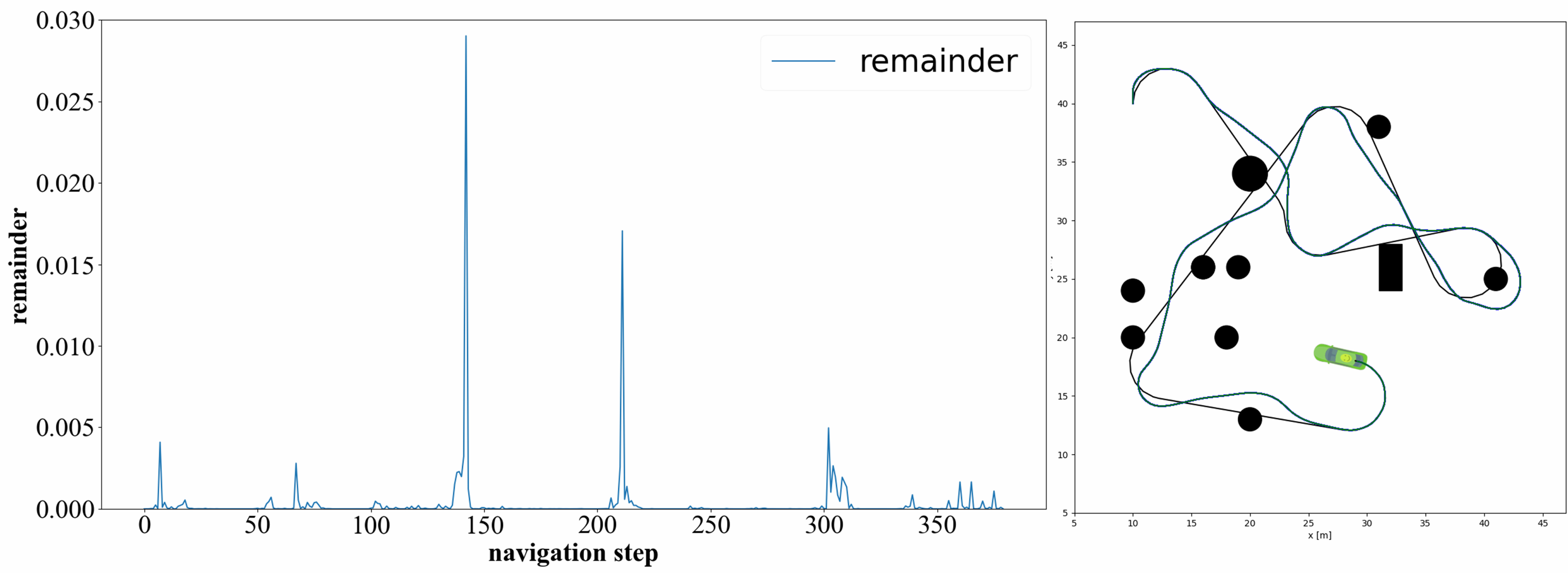}
  \caption{Accuracy of the adopted linear approximation.}
  \label{compare}
\end{figure}

For linear robot dynamics (e.g., omni wheels), we have $f({\mathbf{s}_t},{\mathbf{u}_t}) =
\left[(\mathbf{A}_t-\mathbf{I}){\mathbf{s}_t} + \mathbf{B}_t{\mathbf{u}_t} + \mathbf{c}_t)\right/\Delta t]$, and 
\begin{equation}
  {\mathbf{s}_{t + 1}} = \mathbf{A}_t{\mathbf{s}_t} + \mathbf{B}_t{\mathbf{u}_t} + \mathbf{c}_t,~t=0,\cdots,N-1,
  \label{dynamics}
\end{equation}
where $(\mathbf{A}_t$, $\mathbf{B}_t$, $\mathbf{c}_t)$ are the coefficient matrices.
For nonlinear robot dynamics (e.g., Ackerman wheels~\cite{lynch2017modern}), we have $$f({\mathbf{s}_t},{\mathbf{u}_t}) = \left[
    {{v_t}\cos ({\theta _t})}, 
    {{v_t}\sin ({\theta _t})}, 
    {\frac{{{v_t}\tan {\psi _t}}}{L}}
  \right]^T.$$
By leveraging the first-order Taylor polynomial, the nonlinear dynamics function can be linearized into the same structure as \eqref{dynamics}, where the associated coefficients $(\mathbf{A}_t$, $\mathbf{B}_t$, $\mathbf{c}_t)$ at time $t$ in \eqref{dynamics} are given by:
\begin{align}
    \mathbf{A}_t &= \left[ {\begin{array}{*{20}{c}}
    1&0&{ - \bar{v}_t\sin (\bar{\theta}_t) {\Delta t} }\\
    0&1&{\bar{v}_t\cos (\bar{\theta}_t) {\Delta t}}\\
    0&0&{1}
    \end{array}} \right], 
    \\
    \mathbf{B}_t &= \left[ {\begin{array}{*{20}{c}}
    {\begin{array}{*{20}{c}}
    {\cos (\bar{\theta}_t) {\Delta t} }&0
    \end{array}}\\
    {\begin{array}{*{20}{c}}
    {\sin (\bar{\theta}_t) {\Delta t} }&0
    \end{array}}\\
    {\begin{array}{*{20}{c}}
    {\frac{{\tan \bar{\psi}_t {\Delta t}}}{L}}&{\frac{\bar{v}_t {\Delta t}}{{L{{\cos }^2}\bar{\psi}_t }}}
    \end{array}}
    \end{array}} \right],
    \\ 
    \mathbf{c}_t &= \left[ {\begin{array}{*{20}{c}}
    {\bar{\theta}_t \bar{v}_t\sin (\bar{\theta}_t) {\Delta t} }\\
    { - \bar{\theta}_t \bar{v}_t\cos (\bar{\theta}_t) {\Delta t} }\\
    { - \frac{{\bar{\psi}_t \bar{v}_t {\Delta t}}}{{L{{\cos }^2}\psi_t }}}
    \end{array}} \right],
\end{align}
where $\bar{v}_t$ and $\bar{\psi}_t$ are the nominal linear speed and steering angle of the control vector $\mathbf{\bar{u}}_t$ at time $t$, respectively. The constant $L$ denotes the distance between the front and rear axles.
To evaluate the accuracy of the adopted linerization, the trajectories generated by the original dynamics function (green line) and linearized dynamics function (blue line) are shown in Fig.~\ref{compare}. 
It can be seen that the two trajectories are very close to each other. 
Moreover, as seen in Fig.~\ref{compare}, the approximation error fluctuates between 0 and 0.03, which is negligible.
This is because we can set $\bar{\theta}_t,\bar{v}_t,\bar{\psi}_t$ as the robot states from the last navigation step, and the robot states have a slight change between consecutive steps.

In practice, there are boundary constraints on the control vector $\mathbf{u}_t$:
\begin{equation}
  {\mathbf{u}_{\min }} \preceq \mathbf{u}_t \preceq {\mathbf{u}_{\max }},~{\mathbf{a}_{\min }} \preceq {\mathbf{u}_{t + 1}} - {\mathbf{u}_t} \preceq {\mathbf{a}_{\max }},
  \label{boundary}
\end{equation}
where ${\mathbf{u}_{\min }},{\mathbf{u}_{\max }} \in {\mathbb{R}^{{n_u}}}$ are the minimum and maximum values of the control vector, respectively.
The minimum and maximum values of the acceleration vector are denoted by $\mathbf{a}_{\max }$ and ${\mathbf{a}_{\min }}$, respectively.

\subsection{Problem Formulation}
We consider the MPC formulation for autonomous planning, where the future state is predicted by the robot dynamics within a receding horizon consisting of $N$ time steps. 
The goal of MPC is to minimize a convex and smooth cost function $C_0(\{\mathbf{s}_t,\mathbf{u}_t\}_{\forall t})$ under collision avoidance, state evolution, and boundary constraints. The explicit form of $C_0$ depends on the type of navigation task. 
For instance, the cost function of path tracking is given by
\begin{align}
  &C_0(\{\mathbf{s}_t,\mathbf{u}_t\}) =  \sum\limits_{t = 0}^{N} \left[Q_t{{({\mathbf{s}_t} - {\mathbf{s}_{t}^\diamond})}^2}  + P_t{({v_t} - {v_{t}^\diamond})^2}\right],
  \label{nav_cost}
\end{align}
where ${\mathbf{s}_{t}^\diamond}$ and ${v_{t}^\diamond}$ are the reference state and speed. A larger value of the weighting coefficients $\{Q_t, P_t\}$ leads to the robot moving with the reference path. 

The MPC optimization problem over a finite future horizon ($t=1, \cdots N$) with $M$ obstacles is obtained by combining the constraints in (\ref{cac}), (\ref{dynamics}), (\ref{boundary}):
\begin{subequations} 
    \begin{align}
    \mathcal{P}_0:\min \limits_{\{\mathbf{s}_t, \mathbf{u}_t\}}~~& C_{0}(\{\mathbf{s}_t, \mathbf{u}_t\}) \\
    \text { s.t. }~~& \mathbf{s}_{t+1}=\mathbf{A}_t \mathbf{s}_{t}+\mathbf{B}_t\mathbf{u}_{t}+ \mathbf{c}_t, ~\forall t,\label{problemb}\\
    & \mathbf{u}_{\min } \preceq \mathbf{u}_{t} \preceq \mathbf{u}_{\max },~\forall t, 
    \\
    &\mathbf{a}_{\min }  \preceq  {\mathbf{u}_{t + 1}} - {\mathbf{u}_t}   \preceq \mathbf{a}_{\max },~\forall t, \label{problemc}\\
    & \operatorname{\bf{dist}}\left(\mathbb{Z}_t\left(\mathbf{s}_{t}\right),\mathbb{O}_{m}\right) \geq d_{\mathrm{safe}},~\forall t,m. \label{probleme}
    \end{align}
    \label{problem}
\end{subequations}
The challenges of solving $\mathcal{P}_0$ in cluttered environments are twofold: 1) {\eqref{probleme}} is a sufficient but not necessary condition for collision avoidance, making it difficult to be satisfied; 2) the number of obstacles $M$ in {\eqref{probleme}} could be large. 
Below, an effective method will be presented to tackle these challenges.

\section{RDA Collision-Free Motion Planner}\label{section4} 

In this section, the RDA approach for solving $\mathcal{P}_0$ is presented. The overall algorithm is summarized in Algorithm 1.

\subsection{$l_1$-Regularization} 
To address the first challenge, we need to reformulate the constraint {\eqref{probleme}}. 
Intuitively, an ideal strategy is to generate a larger $d_{\mathrm{safe}}$ for the upcoming time steps and a smaller $d_{\mathrm{safe}}$ for the future time steps. To this end, this paper proposes a dynamic safety distance approach, where $d_{\mathrm{safe}}$ is replaced by a vector variable $\mathbf{d}=[d_1,\cdots,d_N]^T\in{\mathbb{R}^{N}}$, and equation {\eqref{probleme}} is converted into $ \operatorname{\bf{dist}}\left(\mathbb{Z}_t\left(\mathbf{s}_{t}\right),\mathbb{O}_{m}\right) \geq d_t$. Each $d_t$ is upper bounded by $d_{\max }$ and lower bounded by $d_{\min }$. 
However, such a formulation ignores the different importance of upcoming and future states, which leads to a solution with $d_t = d_{\min}$ for all $t$. 
Consequently, the $l_1$-regularization method is further proposed to generate uneven values of $\{d_t\}$ among different time steps by imposing sparsity upon $\mathbf{d}$. 
This would automatically allocate different attention to different states. 
Mathematically, $l_1$-regularization is realized by adding a penalty function of $\mathbf{d}$ to $C_0$. The penalty function is set as the negative $l_1$-norm of $\mathbf{d}$, i.e., ${C_1}(\mathbf{d}) = - \eta \|\mathbf{d}\|_1= - \eta \sum_{t = 0}^N {d_t}$, where $\eta \in \mathbb{R} \ge 0$ is a weight factor to adjust the collision avoidance ability in different scenarios:
\begin{subequations} 
  \begin{align}
  \min \limits_{\{\mathbf{s}_t, \mathbf{u}_t\},\mathbf{d}}~~& C_{0}(\{\mathbf{s}_t, \mathbf{u}_t\}) + C_{1}(\mathbf{d}) \\
  \text { s.t. }~~&\eqref{problemb}-\eqref{problemc},
  \\
  &
  d_t\in[{d_{\min }},{d_{\max }}],~\forall t,
  \\
  & \operatorname{\bf{dist}}\left(\mathbb{Z}_t\left(\mathbf{s}_{t}\right),\mathbb{O}_{m}\right) \geq d_{t},~\forall t,m. \label{probleme_2}
  \end{align}
  \label{problemp1}
\end{subequations}
~Note that a larger value of $\eta$ encourages ${d_t}$ to reach ${d_{\max }}$ and vice versa. 


\subsection{Parallel Computation via ADMM} 
To address the second challenge, we first transform the constraint $ \operatorname{\bf{dist}}\left(\mathbb{Z}_t\left(\mathbf{s}_{t}\right),\mathbb{O}_{m}\right) \geq d_{t}$ into its linearized dual form:
\begin{equation}
  \begin{array}{l}
\operatorname{\bf{dist}}\left(\mathbb{Z}_t\left(\mathbf{s}_{t}\right),\mathbb{O}_{m}\right) \geq d_{t} \Leftrightarrow
\\\bm{\lambda}_{t,m} \succeq_{\mathcal{O}_m^*} \mathbf{0},~ \bm{\mu}_{t,m} { \succeq_{\mathcal{K}_r^*}}\mathbf{0},\\
     \bm{\lambda}_{t,m}^T\mathbf{D}_m\mathbf{p}_t(\mathbf{s}_t) - {\bm{\lambda}_{t,m}^T}\mathbf{b}_m - {\bm{\mu}_{t,m}^T}\mathbf{h} \ge {d}_t,
    \\
    {\bm{\mu}_{t,m}^T}\mathbf{G} + {\bm{\lambda}_{t,m} ^T}\mathbf{D}_m\mathbf{R}_t(\mathbf{s}_t) = 0,\\
    {\left\| {{\mathbf{D}_m^T}\bm{\lambda}_{t,m} } \right\|_*} \le 1,
    \end{array}
    \label{dual}
\end{equation}
where $\bm{\lambda}_{t,m} \in {\mathbb{R}^{l_m}}$ and $\bm{\mu}_{t,m}\in {\mathbb{R}^h}$ are the dual variables. 
The symbol ${\mathcal{K}^*}$ is the dual cone of $\mathcal{K}$ and ${\left\| {\cdot} \right\|_*}$ is the dual norm. 
The dual norm of the Euclidean norm is equal to itself. Note that the above transformation is different from that in{~\cite{zhang2020optimization}}. 
Specifically, the method in{~\cite{zhang2020optimization}} does not involve linearization of translation and rotation matrices, resulting in a smooth nonconvex dual problem. 
In contrast, by leveraging joint linearization and duality transformation, ({\ref{dual}}) has a smooth bi-convex dual form, which facilitates the subsequent parallelization procedure.

Plugging \eqref{dual} into \eqref{problemp1}, 
the problem \eqref{problemp1} is converted into be a bi-convex problem:
\begin{subequations} 
  \begin{align}
  &\min_{\substack{\{\mathbf{s}_t, \mathbf{u}_t,d_t\}
  \\
  \{\bm{\lambda}_{t,m}, \bm{\mu}_{t,m},z_{t,m}\}}}~~
  C_{0}(\{\mathbf{s}_t, \mathbf{u}_t\}) + C_{1}(\mathbf{d})\\
  & \quad\text { s.t. }~~
   \mathbf{s}_{t+1}=\mathbf{A}_t \mathbf{s}_{t}+\mathbf{B}_t  \mathbf{u}_{t}+ \mathbf{c}_t, ~\forall t, \label{admm_1_s}\\
  & \quad\quad\quad\,\,\,\mathbf{u}_{\min } \preceq \mathbf{u}_{t} \preceq \mathbf{u}_{\max },~\forall t, 
  \\
  &\quad\quad\quad\,\,\,\mathbf{a}_{\min }  \preceq \mathbf{a}_{t}  \preceq \mathbf{a}_{\max },~\forall t, \\
  &\quad\quad\quad\,\,\,d_t\in[{d_{\min }},{d_{\max }}],~\forall t,\label{admm_1_e}
  \\
  &\quad\quad\quad\,\,\,z_{t,m}\geq 0,~\forall t,m,\label{admm_2_s}
  \\
  &\quad\quad\quad\,\,\,\bm{\lambda}_{t,m} \succeq_{\mathcal{O}_m^*} \mathbf{0},~ \bm{\mu}_{t,m} { \succeq_{\mathcal{K}_r^*}}\mathbf{0},~\forall t,m,
      \\&\quad\quad\quad\,\,\,
  {\left\| {{\mathbf{D}_m^T}\bm{\lambda}_{t,m} } \right\|_*} \le 1,~\forall t,m, \label{funch}
  \\&\quad\quad\quad\,\,\,
  H_{t,m}(\mathbf{s}_t,\bm{\lambda}_{t,m}, \bm{\mu}_{t,m})=0,~\forall t,m, \label{funci}
  \\&\quad\quad\quad\,\,\,
  I_{t,m}(\mathbf{s}_t,\bm{\lambda}_{t,m}, \bm{\mu}_{t,m},d_t,z_{t,m})=0,~\forall t,m, \label{funcj}
  \end{align}
  \label{problemp3}
\end{subequations}
where, the nonlinear functions in \eqref{funci} and \eqref{funcj} are
\begin{align}
    &H_{t,m}(\mathbf{s}_t,\bm{\lambda}_{t,m}, \bm{\mu}_{t,m})
    = {\bm{\mu}_{t,m}^T}\mathbf{G} + {\bm{\lambda}_{t,m} ^T}\mathbf{D}_m\mathbf{R}_t(\mathbf{s}_t),
\\
&I_{t,m}(\mathbf{s}_t,\bm{\lambda}_{t,m}, \bm{\mu}_{t,m},d_t,z_{t,m})=
\bm{\lambda}_{t,m}^T\mathbf{D}_m\mathbf{p}_t(\mathbf{s}_t) - {\bm{\lambda}_{t,m}^T}\mathbf{b}_m 
\nonumber\\
&\quad\quad\quad\quad\quad\quad\quad\quad\quad\quad\quad\quad
- {\bm{\mu}_{t,m}^T}\mathbf{h} -{d}_t-z_{t,m}.
\end{align}
Existing methods approximate these bi-linear terms by Taylor polynomial~\cite{xia2022trajectory} and solve a sequence of convex problems using the interior point method (IPM). 
The associated computation complexity is $\mathcal{O}((N(n_r+n_u+1)+N(\sum_{m=1}^Ml_m+Mh))^{3.5})$, which is exceedingly high for large $M$. 

In contrast, we solve this bi-convex optimization problem by ADMM, where each iteration includes smaller convex subproblems, and the dual variables associated with different obstacles are updated in parallel to accelerate the computation speed.
In particular, the augmented Lagrangian (scaled form) of \eqref{problemp3} can be formulated as
\begin{align}
    &\mathcal{L}
    \left(
    \{\mathbf{s}_t, \mathbf{u}_t,d_t\},
    \{\bm{\lambda}_{t,m}, \bm{\mu}_{t,m},z_{t,m}\},\{\xi_{t,m},\zeta_{t,m}\}
    \right)
    \nonumber\\
    &=C_{0}(\{\mathbf{s}_t, \mathbf{u}_t\}) + C_{1}(\mathbf{d})
    +J(\{\mathbf{s}_t, \mathbf{u}_t,d_t\})
        \nonumber\\
    &\quad
    +\sum\limits_{t=0}^N \sum\limits_{m=1}^M Q_{t,m}(\bm{\lambda}_{t,m}, \bm{\mu}_{t,m}, z_{t,m})
    \nonumber\\
        &\quad
    +\frac{\rho}{2} \sum_{t=0}^{N} \sum_{m=0}^{M}\left\|I_{t,m}(\mathbf{s}_t,\bm{\lambda}_{t,m}, \bm{\mu}_{t,m},d_t,z_{t,m})
    +\zeta_{t,m}\right\|_{2}^{2} \nonumber\\
    &\quad
    +\frac{\rho}{2} \sum_{t=0}^{N} \sum_{m=0}^{M}\left\|H_{t,m}(\mathbf{s}_t,\bm{\lambda}_{t,m}, \bm{\mu}_{t,m})+\xi_{t,m}\right\|_{2}^{2},
    \label{lang}
\end{align}
where $\{\xi_{t,m},\zeta_{t,m}\}$ are dual variables corresponding to the equality constraints \eqref{funci} and \eqref{funcj}, $\rho$ is the penalty parameter chosen as a large value. 
Here $J(\{\mathbf{s}_t, \mathbf{u}_t,d_t\})$ is the indicator function of the constraints \eqref{admm_1_s}--\eqref{admm_1_e}, i.e., 
$J(\{\mathbf{s}_t, \mathbf{u}_t,d_t\})=0$ if \eqref{admm_1_s}--\eqref{admm_1_e} are satisfied and $J(\{\mathbf{s}_t, \mathbf{u}_t,d_t\})=\infty$ otherwise.
Similarly, $Q_{t,m}(\bm{\lambda}_{t,m}, \bm{\mu}_{t,m}, z_{t,m})$ is the indicator function of the $(t,m)$th constraints in \eqref{admm_2_s}--\eqref{funch}.

In \eqref{lang}, the first three items are coupled across different $t$ due to the constraint \eqref{admm_1_s}, while the latter three terms are decomposable w.r.t. $t$ and $m$. Consequently, we split the primal variables into two groups: 
1) $\{\mathbf{s}_t, \mathbf{u}_t,d_t\}$; 2) $\{\bm{\lambda}_{t,m}, \bm{\mu}_{t,m},z_{t,m}\}$. 
Given the fact that the dual variables $\{\xi_{t,m},\zeta_{t,m}\}$ can be updated in parallel, the ADMM method for minimizing the augmented Lagrangian is 
\begin{subequations}
  \begin{align}
  &\{\mathbf{s}_{t}^{k+1}, \mathbf{u}_{t}^{k+1}, d_{t}^{k+1}\}=\underset{\{\mathbf{s}_t, \mathbf{u}_t,d_t\}}{\arg \min }
  \mathcal{L}
    \Big(
    \{\mathbf{s}_t, \mathbf{u}_t,d_t\},
    \nonumber\\
    &\quad\quad\quad\quad\quad\quad
    \{\bm{\lambda}_{t,m}^{k}, \bm{\mu}_{t,m}^{k},z_{t,m}^{k}\},\{\xi_{t,m}^{k},\zeta_{t,m}^{k}\}
    \Big), \label{admmupdate1}
  \\
  &
  \bm{\lambda}_{t,m}^{k+1},\bm{\mu}_{t,m}^{k+1},z_{t,m}^{k+1}
  =\underset{\bm{\lambda}_{t,m}\bm{\mu}_{t,m},z_{t,m}}{\arg \min }
  \mathcal{L}
    \Big(
    \{\mathbf{s}_t^k, \mathbf{u}_t^k,d_t^k\},
    \nonumber\\
    &\quad\quad\quad\quad
    \{\bm{\lambda}_{t,m}, \bm{\mu}_{t,m},z_{t,m}\},\{\xi_{t,m}^{k},\zeta_{t,m}^{k}\}
    \Big), \forall t,m,  \label{admmupdate2}\\
  &\xi_{t,m}^{k+1}=\xi_{t,m}^{k}
+{ ({\bm{\mu}_{t,m}^k})^T}\mathbf{G} + {({\bm{\lambda}_{t,m}^k})^T}\mathbf{D}_m\mathbf{R}_t(\mathbf{s}_t^k), \label{admmupdate3}
\\
  &\zeta_{t,m}^{k+1}=\zeta_{t,m}^{k}+
  ({\bm{\lambda}_{t,m}^k})^T\mathbf{D}_m\mathbf{p}_t(\mathbf{s}_t^k) - ({\bm{\lambda}_{t,m}^k})^T\mathbf{b}_m \nonumber 
  \\
  & \qquad \qquad \qquad - ({\bm{\mu}_{t,m}^k})^T\mathbf{h} -d_t^k-z_{t,m}^k, \forall t,m, \label{admmupdate4}
  \end{align}
  \label{admmupdate}
 \end{subequations}
Problems {\eqref{admmupdate1}} and {\eqref{admmupdate2}} are convex, and {\eqref{admmupdate3}} and {\eqref{admmupdate4}} are the subgradients to update the dual variables. Problem \eqref{admmupdate1} can be efficiently solved via a convex solver, e.g. CVXPY~\cite{diamond2016cvxpy}, whose computation cost is independent of $M$. Problems \eqref{admmupdate2}--\eqref{admmupdate4} are all separable across $(t,m)$, and $\{\bm{\lambda}_{t,m}, \bm{\mu}_{t,m},z_{t,m},\xi_{t,m},\zeta_{t,m}\}$ can all be updated in parallel. 
Furthermore, each problem in \eqref{admmupdate2} is a conic constrained least squares problem which can be solved with low-cost solvers. The stopping criteria to terminate the iterative procedure is given by 
 \begin{align}
  &\sum_{t}\sum_m\Big[\|H_{t,m}(\mathbf{s}_t^k,\bm{\lambda}_{t,m}^k, \bm{\mu}_{t,m}^k)\|_2^2
  \nonumber\\
  &
  +\| I_{t,m}(\mathbf{s}_t^k,\bm{\lambda}_{t,m}^k, \bm{\mu}_{t,m}^k,d_t^k,z_{t,m}^k)\|_2^2\Big]
  \leq \epsilon^{\text {pri}}, \label{pcondition}
  \\
       &{\sum_{t}\sum_m\Big[\|
       \bm{\lambda}_{t,m}^{k+1}-\bm{\lambda}_{t,m}^{k}\|_2^2
  +\| \bm{\mu}_{t,m}^{k+1}-\bm{\mu}_{t,m}^{k}\|_2^2\Big]
  \leq \epsilon^{\text {dual}},} \label{dcondition}
\end{align}
where the first condition guarantees the primal residual being small and the second condition guarantees the dual residual being small. Constants $\epsilon^{\text {pri }} > 0$ and $\epsilon^{\text {dual }} > 0$ are the stopping criterion values. 
To achieve the best tradeoff between  the solution quality and computational cost, $\epsilon^{\text {pri}}$ and $\epsilon^{\text {dual}}$ can be chosen empirically.
According to our experimental data in appendix, the stopping criteria value can be set to $\epsilon^{\text {pri}},\epsilon^{\text {dual}} \in [0.1, 1]$.
 
\subsection{Complexity Analysis and Warm Starting}

\emph{1) Complexity Analysis}: 
The entire procedure of RDA is summarized in Algorithm 1.
In each iteration, the CVXPY solver is first adopted to solve \eqref{admmupdate1} with a complexity of $\mathcal{O}((N(n_r+n_u+1))^{3.5})$.
Then, to optimize dual collision variables, we need to solve \eqref{admmupdate2}, which involves $NM$ subproblems with the $m$th problem being solved at a computational cost of $\mathcal{O}((l_m+h+1)^{3.5})$.
Finally, the Lagrange multipliers are updated via vector-matrix multiplications in \eqref{admmupdate3}--\eqref{admmupdate4}, which involve a complexity of $\mathcal{O}(N(\sum_{m-1}^Ml_mn_m+Mhn_r))$. 
Therefore, the total complexity of RDA is given by 
$\mathcal{O}(K(N(n_r+n_u+1))^{3.5}+
N\sum_{m=1}^M(l_m+h+1)^{3.5}+N(\sum_{m-1}^Ml_mn_m+Mhn_r))$, where $K$ is the number iterations for the RDA to converge. 

\emph{2) Warm Starting}: 
It can be seen from the above analysis that the complexity of RDA is linear in $M$. 
This significantly saves computation time compared with IPM whose complexity is cubic in $M$. 
Furthermore, the value of $K$ can be significantly reduced by leveraging warm starting. 
Specifically, for collision-avoidance autonomous driving, the odometry of the ego-vehicle and the states of surrounding obstacles do not vary significantly between two consecutive time steps, and therefore, the current RDA solutions can be adopted as an initial guess of both the primal and dual solutions to the subsequent RDA iterative procedure. 
That is, setting 
$\{\bm{\lambda}_{t,m}^{0}, \bm{\mu}_{t,m}^{0},z_{t,m}^{0},\xi_{t,m}^{0},\zeta_{t,m}^{0}\}
=
\{\bm{\lambda}_{t,m}^{\diamond}, \bm{\mu}_{t,m}^{\diamond},z_{t,m}^{\diamond},\xi_{t,m}^{\diamond},\zeta_{t,m}^{\diamond}\}
$ for \eqref{admmupdate1}, where the solutions with $\diamond$ denote the cached values obtained from the previous RDA outputs.

\section{ Experiments and Results}\label{section5}
In this section, we adopt numerical simulations, high-fidelity simulators, and hardware experiments to verify the performance and efficiency of RDA. In particular, we consider the collision-free path tracking task for autonomous vehicles. The task is defined as moving as close as possible to a reference path while avoiding collisions.
In our experiments, we set $\eta=10$, $\rho=10$.

Besides the proposed RDA planner, we also simulate the following schemes for comparison: (1) \textbf{Optimization based collision avoidance (OBCA) planner}~\cite{zhang2020optimization}, which directly solves the dual-MPC problem via nonconvex optimization methods such as sequential convex programming; (2) \textbf{Time elastic band (TEB) planner}~\cite{rosmann2017kinodynamic}, which obtains a sequence of robot poses via general graph optimization and sparsity regularization; (3) \textbf{Point-mass ADMM (PMA)}~\cite{cheng2021admm}, which models each obstacle as a single point and determines the collision condition by computing the distance between this point and the ego-robot position; (4) \textbf{OBCA-II}, which adopts the OBCA planner but with no safe-distance regularization; (5) \textbf{RDA-II}, which adopts the RDA planner but with no safe-distance regularization.

\begin{algorithm}[t]
  \caption{RDA motion planner}
  \label{rda algorithm}
  Initialize the given points of the robot state ${\mathbf{s}}$ and control vector ${\mathbf{u}}$\;
  \For{iteration $k=1, 2,\cdots$}
  {
    Update the variables $\mathbf{s}$, $\mathbf{u}$, $\mathbf{d}$ by solving {\eqref{admmupdate1}} with CVXPY\;
    
    Update the dual collision variables $\bm{\lambda}$, $\bm{\mu}$, $\mathbf{z}$ by solving {\eqref{admmupdate2}} with accelerated gradient projection in a parallel manner\;
    
    Update the Lagrangian multipliers $\bm{\zeta}$ and $\bm{\xi}$ by \eqref{admmupdate3}-\eqref{admmupdate4} in a parallel manner\;
    \If{\eqref{pcondition} and \eqref{dcondition} are satisfied}{break}
 }
Apply the first receding step control vector to the robot.
\end{algorithm}

\subsection{Numerical Simulation}

\begin{figure*}[t]
  \centering
  \begin{subfigure}[t]{0.24\textwidth}
      \includegraphics[width=0.9\textwidth]{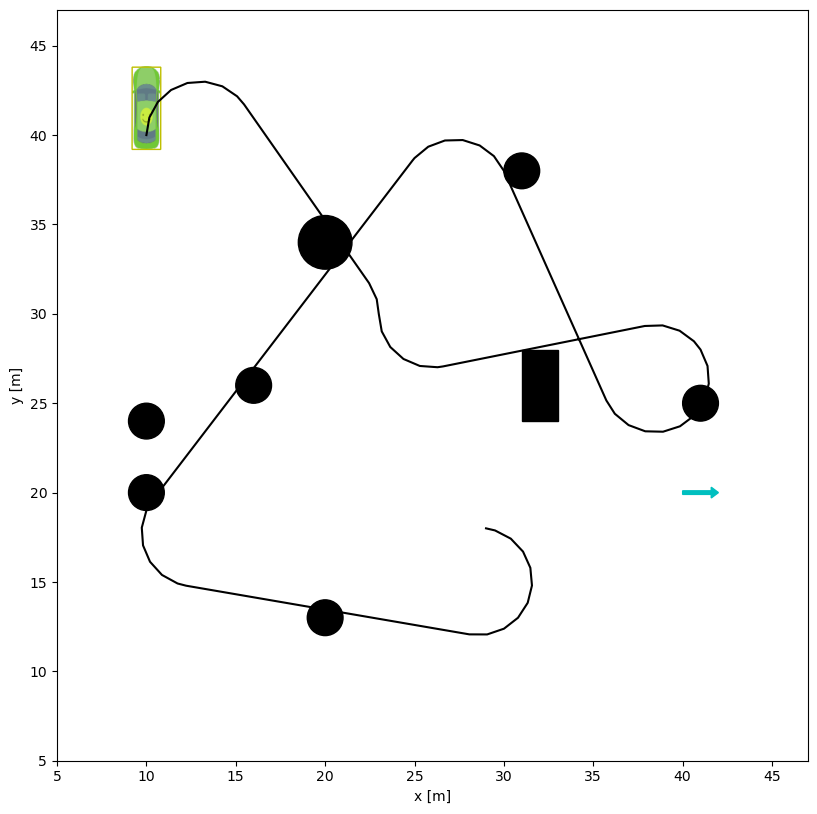}
      \caption{RDA at time step 0}
  \end{subfigure}
  \hfill
  \begin{subfigure}[t]{0.24\textwidth}
    \includegraphics[width=0.9\textwidth]{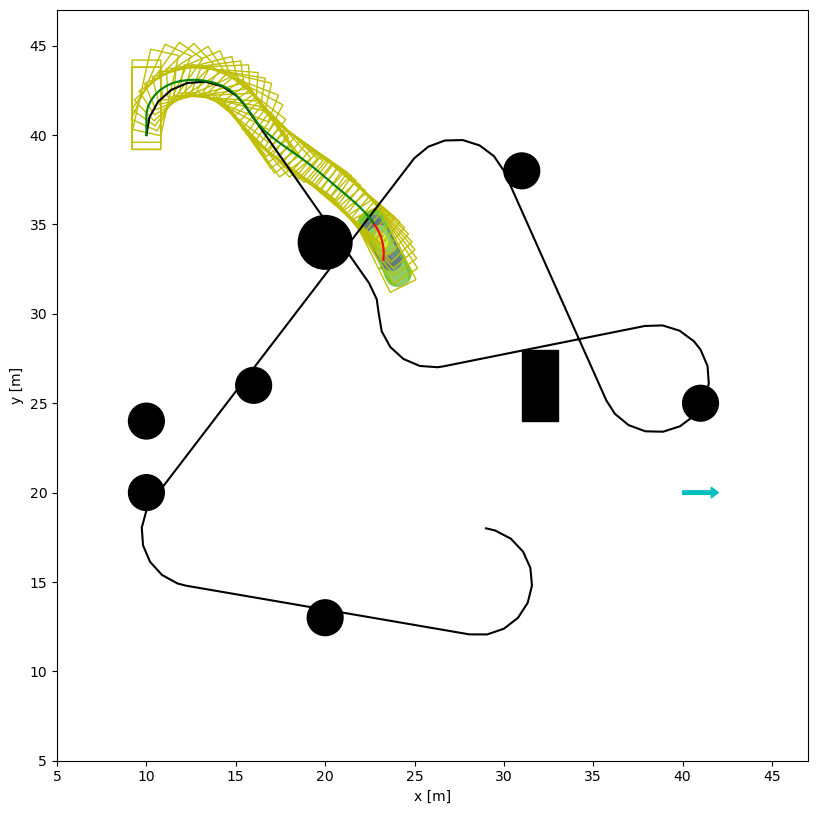}
    \caption{RDA at time step 50}
  \end{subfigure}
  \hfill
  \begin{subfigure}[t]{0.24\textwidth}
    \includegraphics[width=0.9\textwidth]{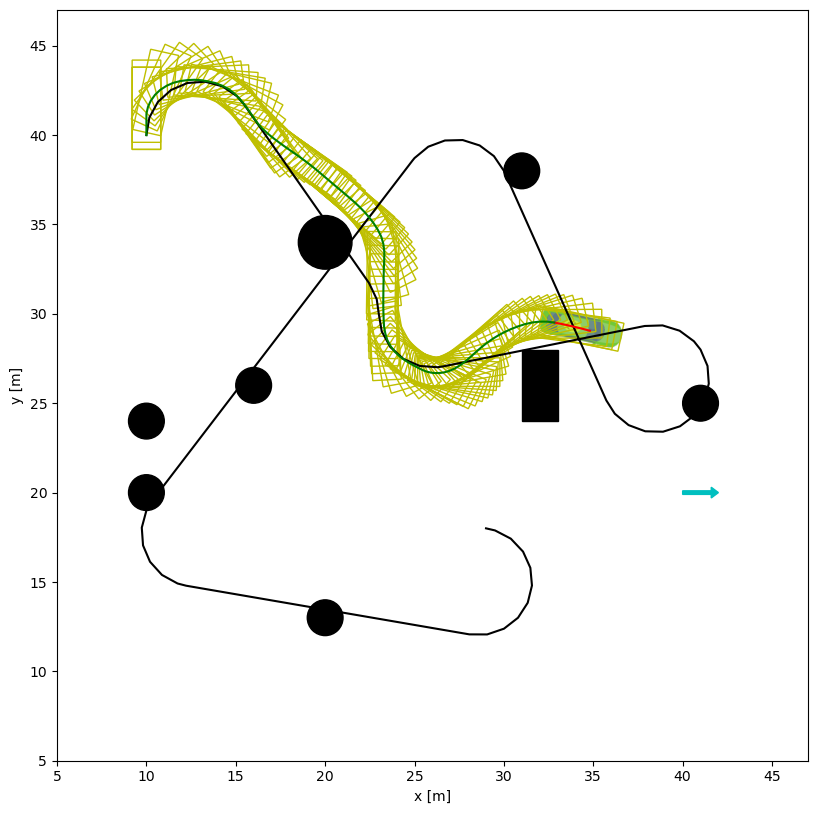}
    \caption{RDA at time step 100}
  \end{subfigure}
  \hfill
  \begin{subfigure}[t]{0.24\textwidth}
    \includegraphics[width=0.9\textwidth]{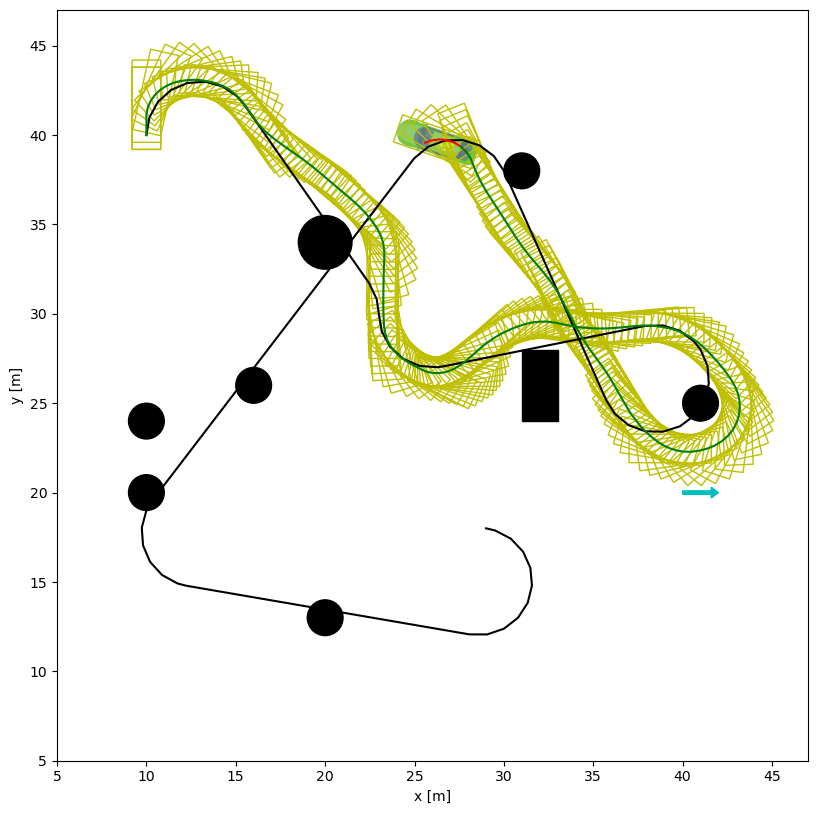}
    \caption{RDA at time step 200}
  \end{subfigure}

  \begin{subfigure}[t]{0.24\textwidth}
    \includegraphics[width=0.9\textwidth]{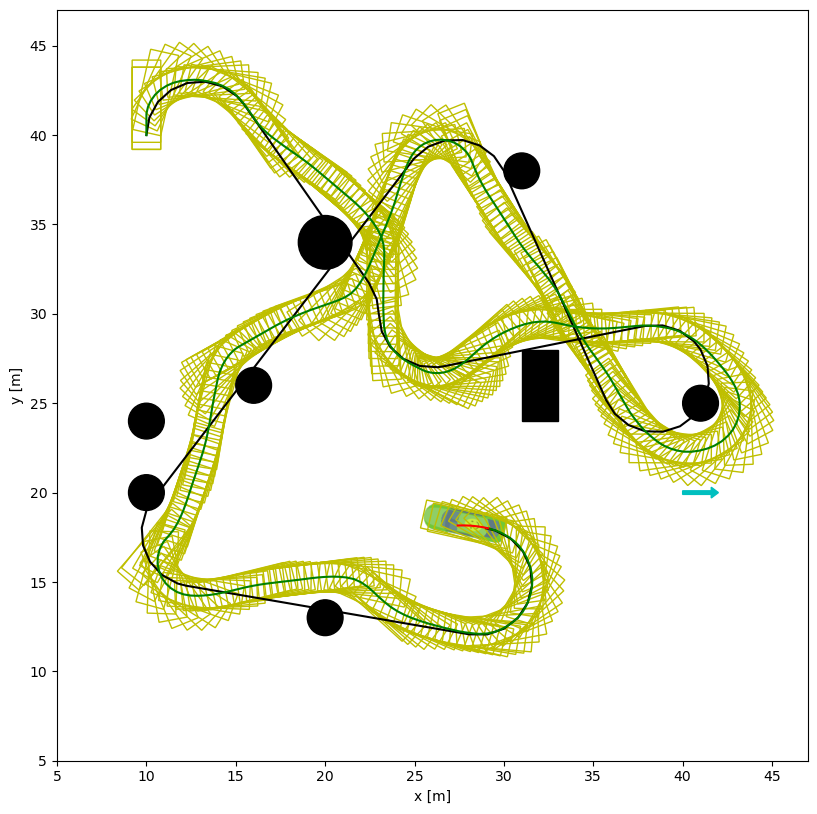}
    \caption{RDA trajectory }
  \end{subfigure}
  \hfill
  \begin{subfigure}[t]{0.24\textwidth}
      \includegraphics[width=0.9\textwidth]{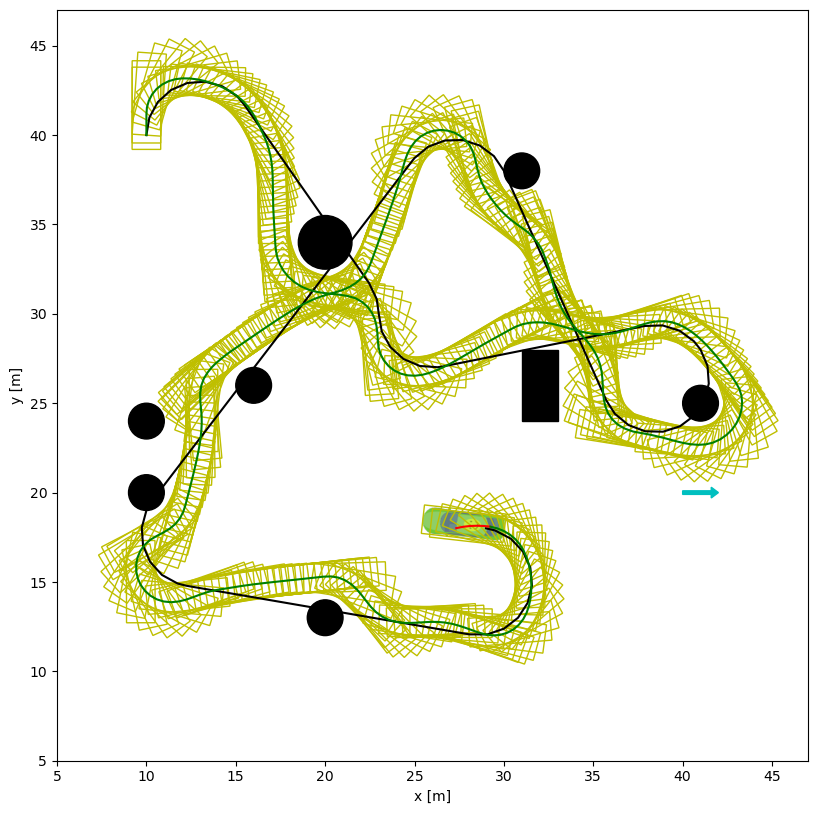}
      \caption{OBCA trajectory}
  \end{subfigure}
  \hfill
  \begin{subfigure}[t]{0.24\textwidth}
    \includegraphics[width=0.9\textwidth]{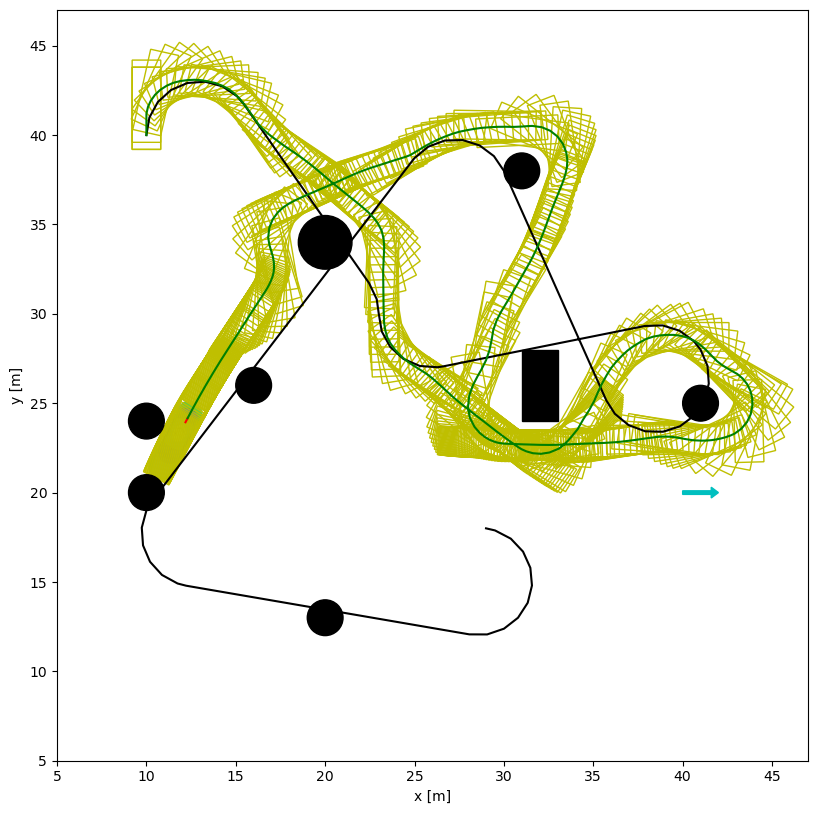}
    \caption{RDA-II trajectory}
  \end{subfigure}
  \hfill
  \begin{subfigure}[t]{0.24\textwidth}
    \includegraphics[width=0.9\textwidth]{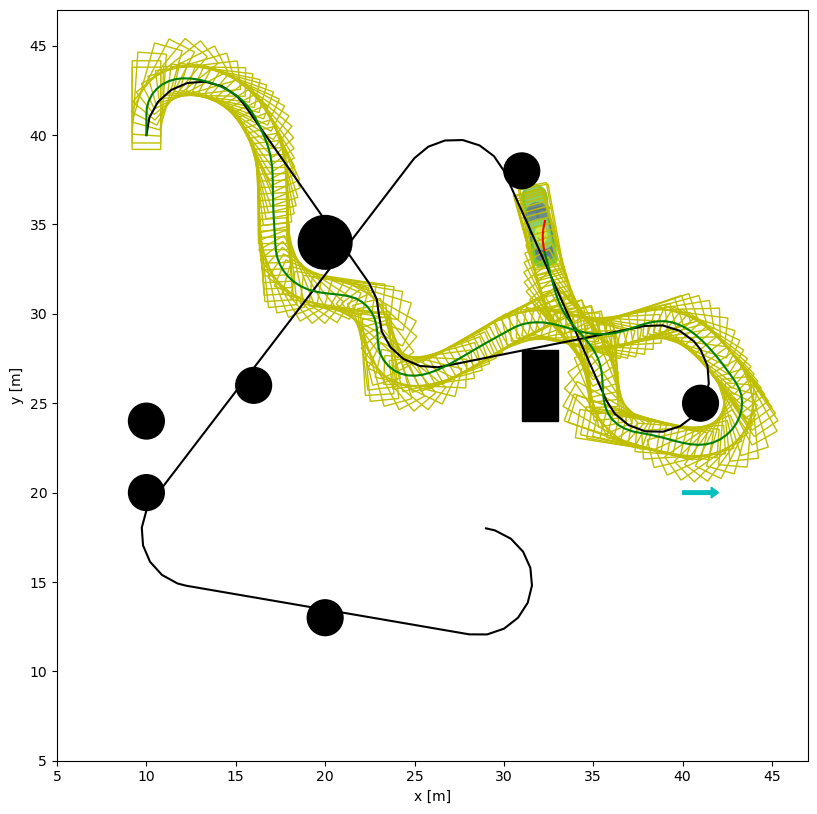}
    \caption{OBCA-II trajectory}
  \end{subfigure}
  \caption{Navigation process and trajectory comparison of RDA and OBCA in a simulated environment.}
  \label{sim1}
\end{figure*}

\begin{figure}[t]
  \centering
    \includegraphics[width=0.45\textwidth]{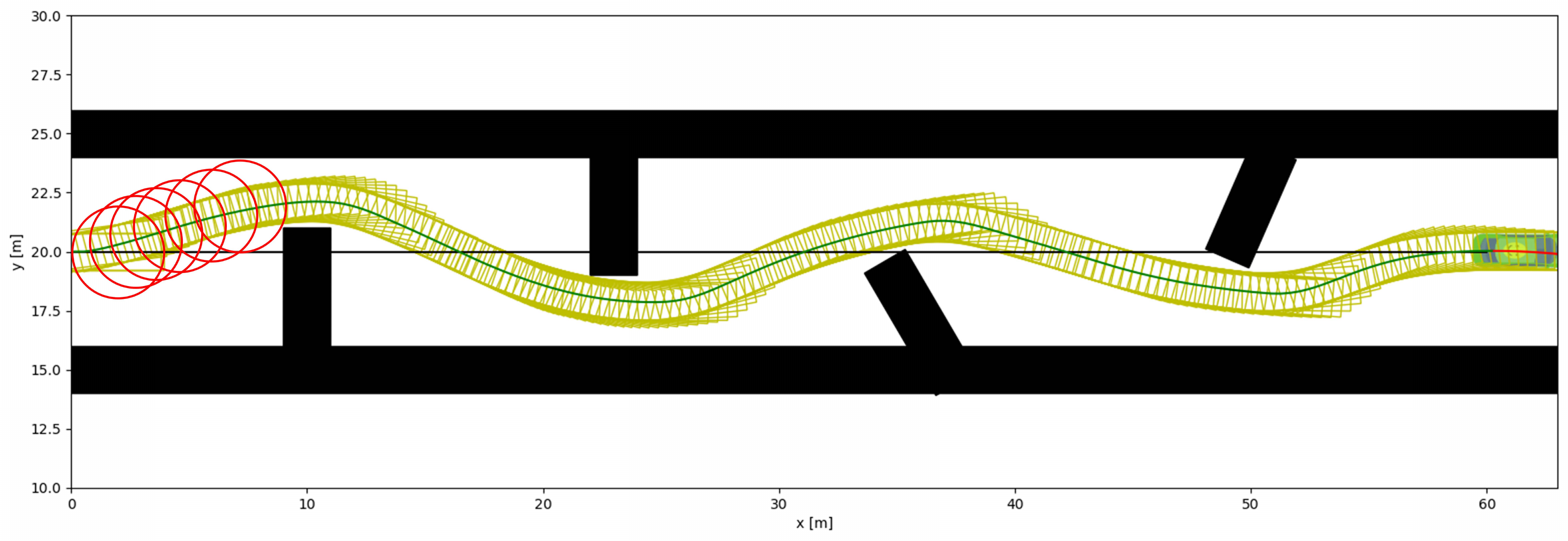}
    \caption{RDA (yellow box) and PMA (red circle) trajectories in the corridor.}
  \label{corridor}
\end{figure}

We verify the proposed RDA in our intelligent robot simulator (\textit{ir-sim})\footnote{\href{https://github.com/hanruihua/ir_sim}{\url{https://github.com/hanruihua/ir_sim}}.}, which is a Python-based 2D numerical simulator for robotic localization and navigation. 
The simulation scenario is given in Fig.~\ref{sim1}, where the obstacles are marked in black with various shapes (e.g., rectangle, circle, etc.).
The reference path (i.e., the black line) in \textit{ir-sim} is generated from a series of directional waypoints and calculated by Dubins path~\cite{shkel2001classification}.

The resulting navigation process of RDA in a cluttered environment with 8 obstacles is shown in Fig. \ref{sim1}(a)--(e). 
The actual robot path is represented by a green line and the robot is marked as a yellow box. The predicted trajectory within the receding horizon is represented by a red curve.
It can be seen that the vehicle successfully avoids all obstacles while moving close to the reference path in a smooth manner. 

\begin{figure}[t]
  \centering
  \begin{subfigure}[t]{0.23\textwidth}
      \includegraphics[width=0.95\textwidth]{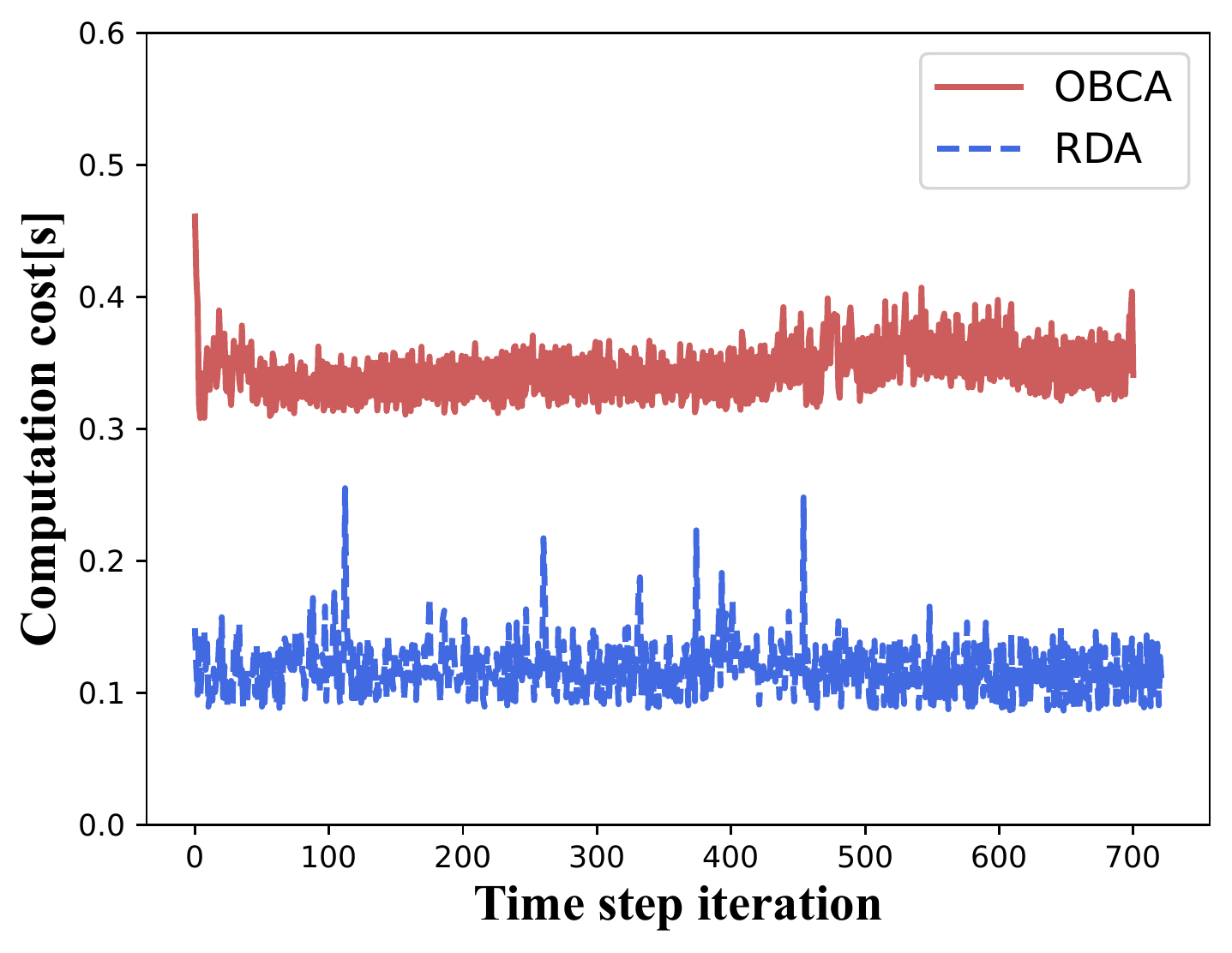}
      \caption{Computation cost in one trial}
  \end{subfigure}
  \begin{subfigure}[t]{0.23\textwidth}
      \includegraphics[width=0.95\textwidth]{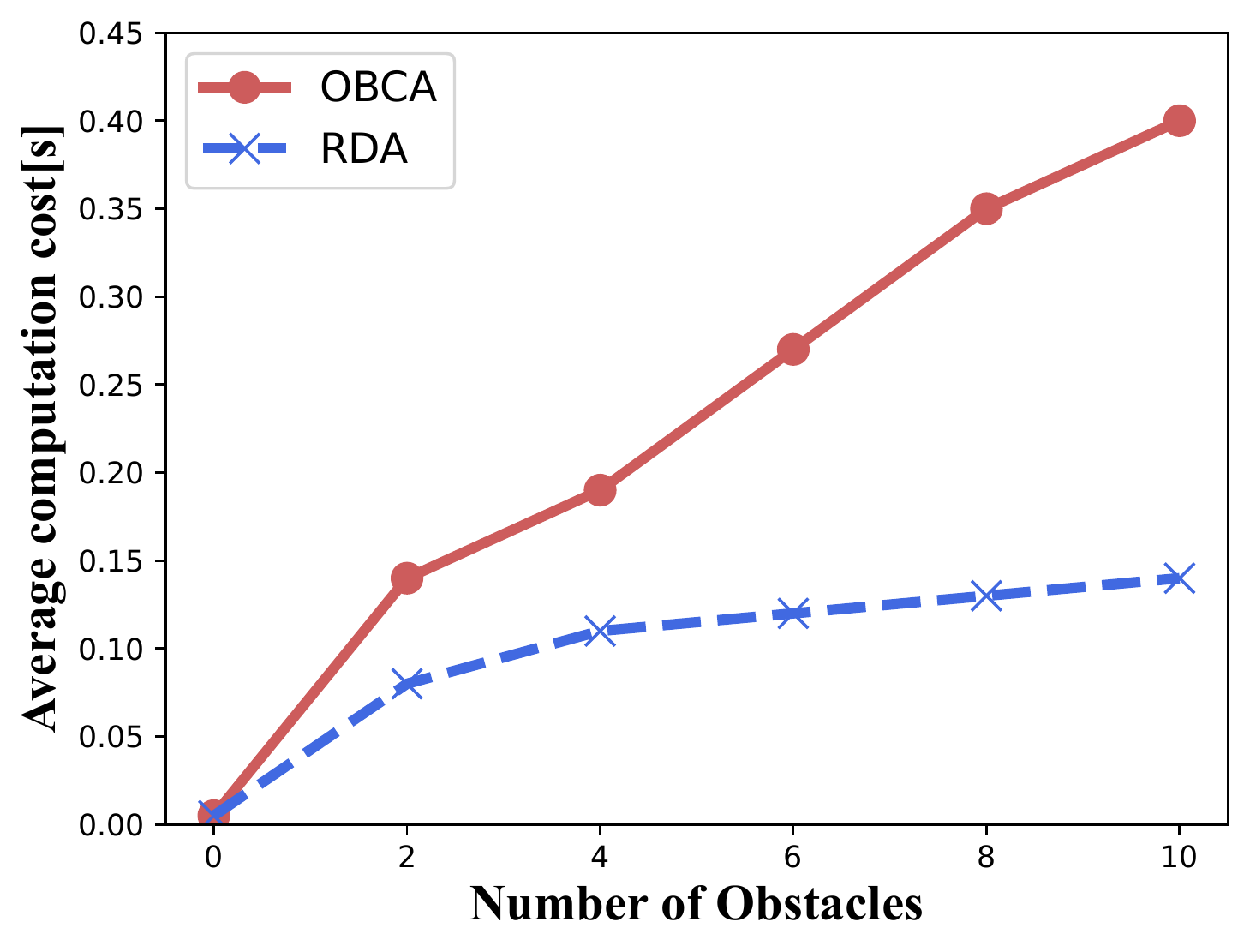}
      \caption{Average computation cost}
  \end{subfigure}
  \caption{Comparison of computation cost of RDA and OBCA.}
  \label{sim2}
\end{figure}

We compare the proposed RDA with OBCA, whose trajectories and computation times are shown in Fig.~\ref{sim1}(e)--(f) and Fig.~\ref{sim2}(a)--(b), respectively. 
It can be seen that both approaches have the ability to find a proper collision-free trajectory. 
However, the required computation time of RDA is significantly shorter than that of OBCA, i.e., 0.1 seconds for RDA versus over 0.3 seconds for OBCA, as illustrated in Fig.~\ref{sim2}(a).
This implies that the planning frequency of RDA is three times as fast as that of OBCA, thus providing real-time potential in high-speed scenarios. 
Furthermore, as shown in Fig.~\ref{sim2}(b), the computation time of RDA only has a slight change as the number of obstacles increases, while that of OCBA grows dramatically. 
This corroborates the parallel computation capability of RDA.

Here, we compare RDA with RDA-II and OBCA-II with the safe distance $d_{\mathrm{safe}}$ being fixed to $0.5\,$m. It can be seen from Fig.~\ref{sim1}(g)--(h) that the vehicle gets stuck at position $(10,20)$ for RDA-II due to the dense obstacles residing within this local region and position $(31,32)$ for OBCA-II due to the 180-degree turn at the right hand side of the reference path. In contrast, the proposed RDA with distance regularization $\eta\in[5, 30]$ passes these challenging points.

Finally, we compare RDA with PMA in another challenging scenario, i.e.,  
the corridor scenario as shown in Fig.~\ref{corridor}. It can be seen that the vehicle with RDA (yellow box) successfully crosses the entire tunnel while that with PMA (red circle) gets stuck at the first narrow gap. This is because PMA computes the distance between the centers of the ego-vehicle and the nearby obstacle, which works well only when both objects are circular. However, in the considered scenario, both the vehicle and the obstacle are rectangles, making the solution to PMA infeasible. 

\subsection{Gazebo Simulation}

\begin{figure*}[t]
  \centering
    \includegraphics[width=0.6\textwidth]{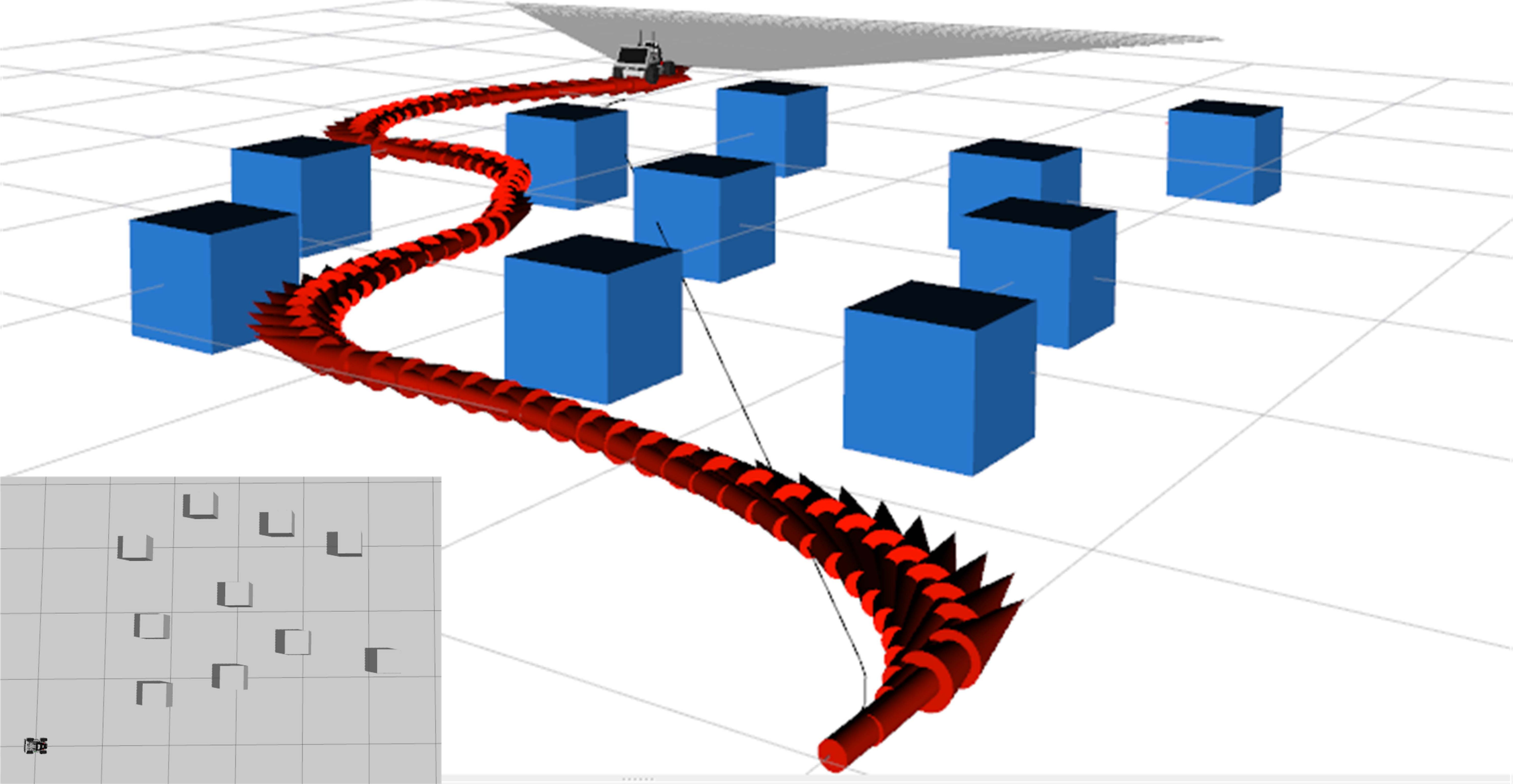}
    \caption{RDA Simulation in the Gazebo.}
  \label{gazebo}
\end{figure*}

To verify the effectiveness of the proposed RDA under practical sensor and motion constraints, we implement the RDA and TEB methods in Gazebo~\cite{koenig2004design}, which is a high-fidelity robotic simulator with close-to-reality sensory data and motion dynamics. 
The scenario consists of an autonomous vehicle and $10$ obstacles (i.e., marked as blue cubes with the size of $0.35\,\mathrm{m}\times0.35\,\mathrm{m}\times0.35\,\mathrm{m}$).
The receding horizon of RDA is set to $20$.
Under the above setting, we execute 50 trials with random goal points and positions of obstacles. 
Fig.~\ref{gazebo} illustrates the trajectory of RDA in one particular case when the robot moves from the starting point $(0, 0)$ to the target point $(5, 4)$.
The odometry of the vehicle with the RDA planner is illustrated by the red arrows. 
It can be seen that the proposed RDA generates a smooth and collision-free trajectory in the dense-obstacle scenario.   

\begin{figure}[t]
  \centering
  \begin{subfigure}[t]{0.23\textwidth}
      \includegraphics[width=0.95\textwidth]{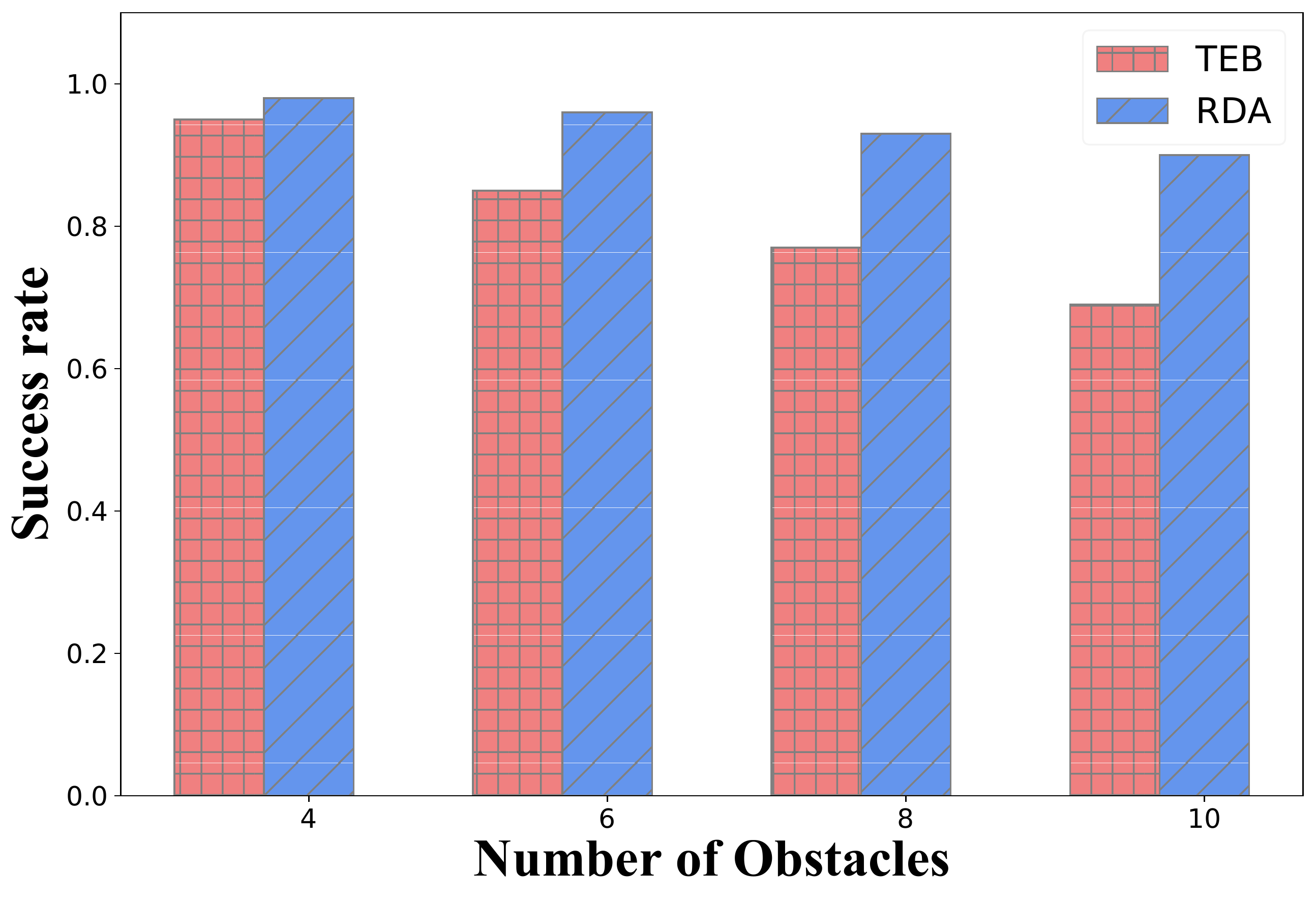}
      \caption{Success rate}
  \end{subfigure}
  \begin{subfigure}[t]{0.23\textwidth}
      \includegraphics[width=0.95\textwidth]{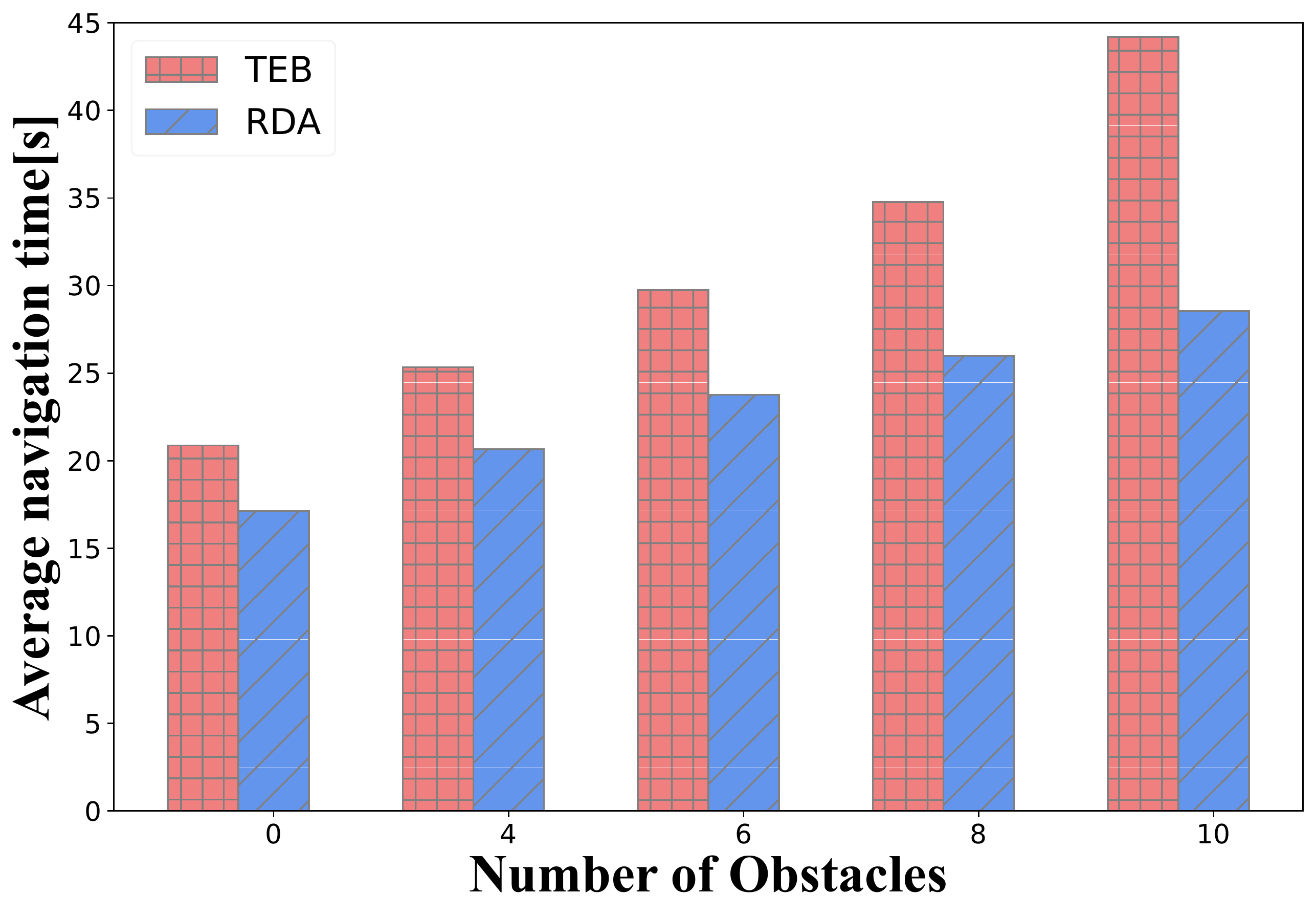}
      \caption{Average navigation time}
  \end{subfigure}
  \caption{Performance comparison of RDA and TEB.}
  \label{sim3}
\end{figure}

Performance comparison between RDA and TEB is shown in Fig.~\ref{sim3}.
The success rates (a successful trail is defined as no collision or stuck during the navigation) of both schemes under the different number of obstacles are shown in Fig.~\ref{sim3}(a).
The two simulated schemes achieve similar success rates in environments with few obstacles. 
However, RDA significantly outperforms TEB in environments with dense obstacles. 
The average navigation time (i.e., the average time required by the planner to move from the starting point to the goal) under the different number of obstacles is shown in Fig.~\ref{sim3}(b). 
RDA reduces the navigation time by up to $40$\%, which demonstrates the necessity of predictive path optimization. 

\begin{figure}[t]
  \centering
    \includegraphics[width=0.45\textwidth]{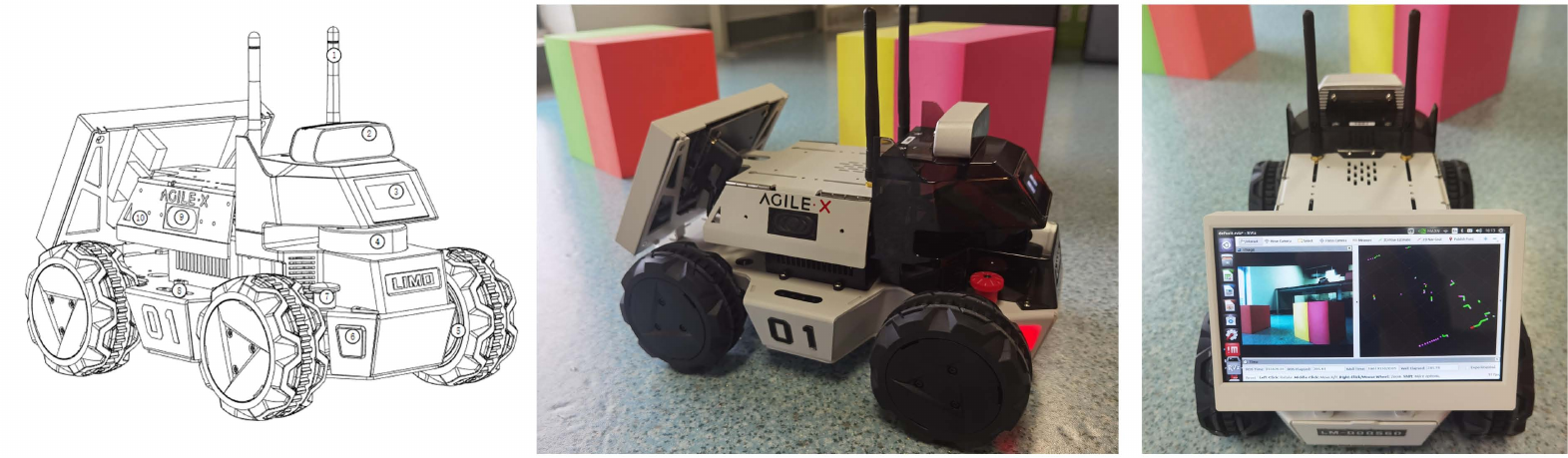}
    \caption{The hardware of the Ackermann robot.}
  \label{hardware}
\end{figure}

\subsection{Hardware Experiment}
\begin{figure*}[t]
  \centering
  \begin{subfigure}[t]{0.32\textwidth}
    \centering
    \includegraphics[width=0.96\textwidth]{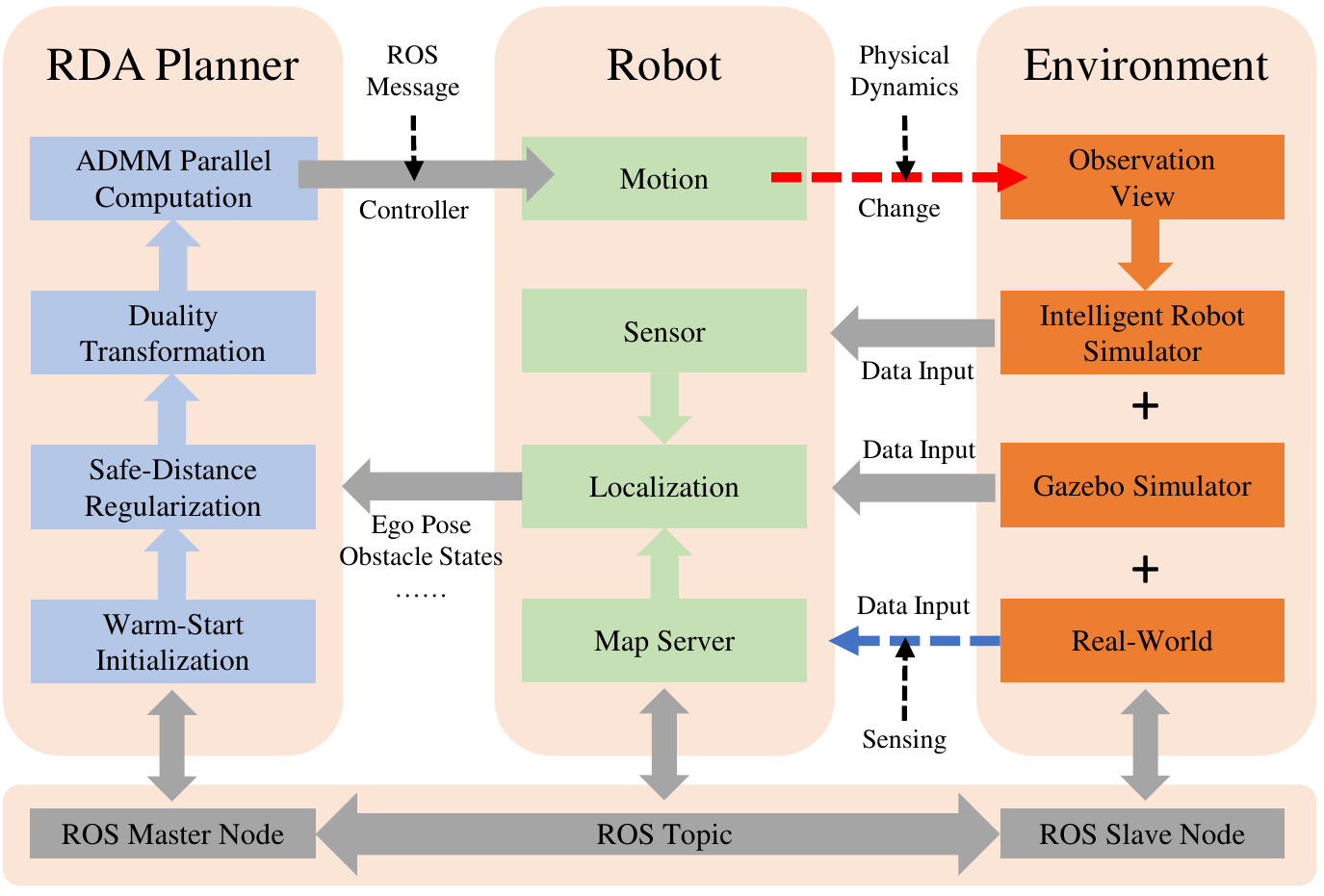}
    \caption{The framework}
    \label{expa}
\end{subfigure}
  \begin{subfigure}[t]{0.32\textwidth}
      \centering
      \includegraphics[width=0.96\textwidth]{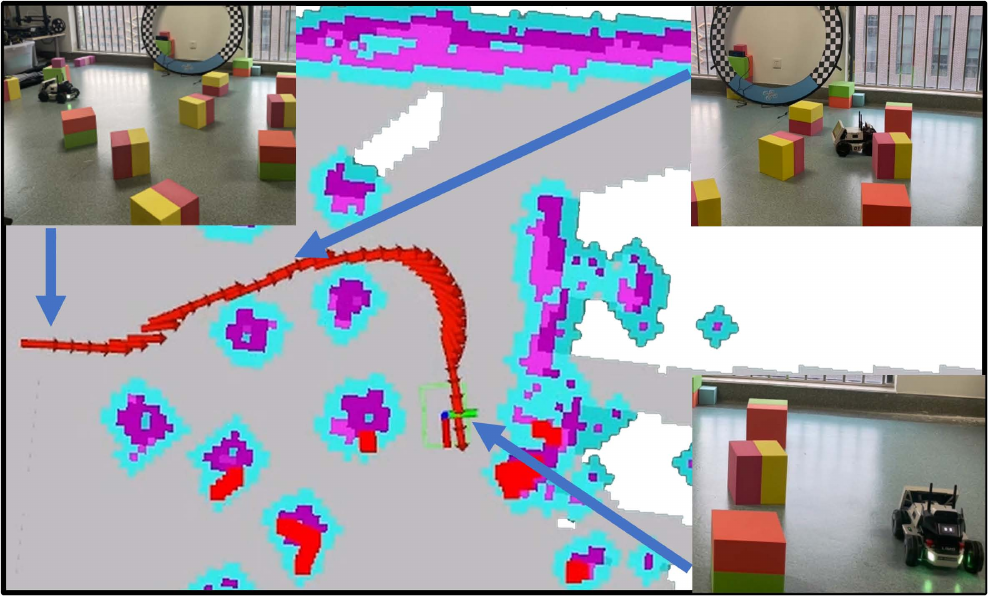}
      \caption{RDA trajectory }
      \label{expb}
  \end{subfigure}
  \begin{subfigure}[t]{0.32\textwidth}
      \centering
      \includegraphics[width=0.96\textwidth]{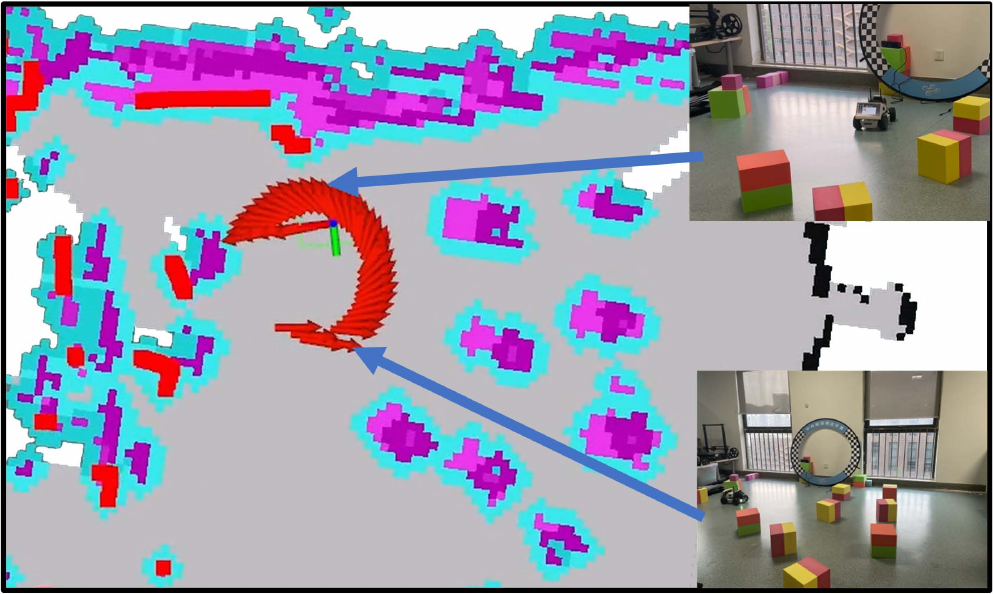}
      \caption{TEB trajectory}
      \label{expc}
  \end{subfigure}
  \caption{Framework and the trajectory comparison of RDA and TEB in a real-world experiment.}
  \label{exp}
\end{figure*}

To evaluate the performance of RDA in real-time operating systems under limited hardware resources, we implement RDA and TEB planners in an autonomous robot shown in Fig.~\ref{hardware}. 
The robot has four wheels and adopts Ackermann steering, which can be viewed as a size-scaled road vehicle. Its sensor suite consists of a 2D Lidar and multiple cameras (including RGB and depth cameras). 
NVIDIA Jetson Nano is adopted as the onboard computing platform for processing the sensor data and executing the navigation package. 
The software architecture of our implementation is shown in Fig.~{{\ref{exp}(\subref{expa})}}, which consists of the RDA planner module and the robot module.
Data sharing between two modules is realized via ROS communications, where master/slave nodes publish/subscribe topics carrying the controller, map, and odometry information.
Note that we convert the obstacles in the costmap into non-point mass representation via DBSCAN~\cite{ester1996density}, and the experimental setting is the same as Gazebo.

Based on the above implementation, the trajectories (red arrows) of RDA and TEB planners of one trial are shown in Fig.~{{\ref{exp}(\subref{expb})}} and Fig.~{{\ref{exp}(\subref{expc})}}, respectively. 
It can be seen that the robot powered by RDA achieves the target point successfully with a smooth curve.
In contrast, the TEB planner fails to find a feasible route to cross those narrow gaps, and the robot eventually gets stuck after a sequence of replanning. 
This demonstrates the effectiveness and robustness of RDA in real time systems and dense obstacle environments. 
Note that we adopt the default parameters for TEB. 
One may fine-tune these parameters to generate a feasible path.
However, TEB involves large-scale parameters without interpretability, leading to a large amount of manual effort for calibration whenever the robot enters a new complex scenario.
In contrast, RDA only involves several tuning parameters (e.g., $\eta$ for regularization and $\rho$ for ADMM augmentation), which significantly facilitates practical deployment. 


\section{Conclusion}\label{section6}
In this paper, we have presented RDA, an accelerated optimization based collision-free motion planner for autonomous navigation. RDA can support Ackermann dynamics and enable adaptive safety distances, non-point-mass representation of obstacles, and parallel computing of optimal trajectories. Various simulations and real-world experiments have shown that the proposed RDA can achieve a significantly shorter execution time compared to existing OBCA; the failure rate and the navigation time of RDA can be reduced by $10\sim40 \%$ compared with various benchmark schemes such as TEB and PMA. Future directions include high-speed vehicle and multi-vehicle RDA, as well as real-world verification.

\appendix

\begin{figure}[!t]
  \centering
  \includegraphics[width=0.48\textwidth]{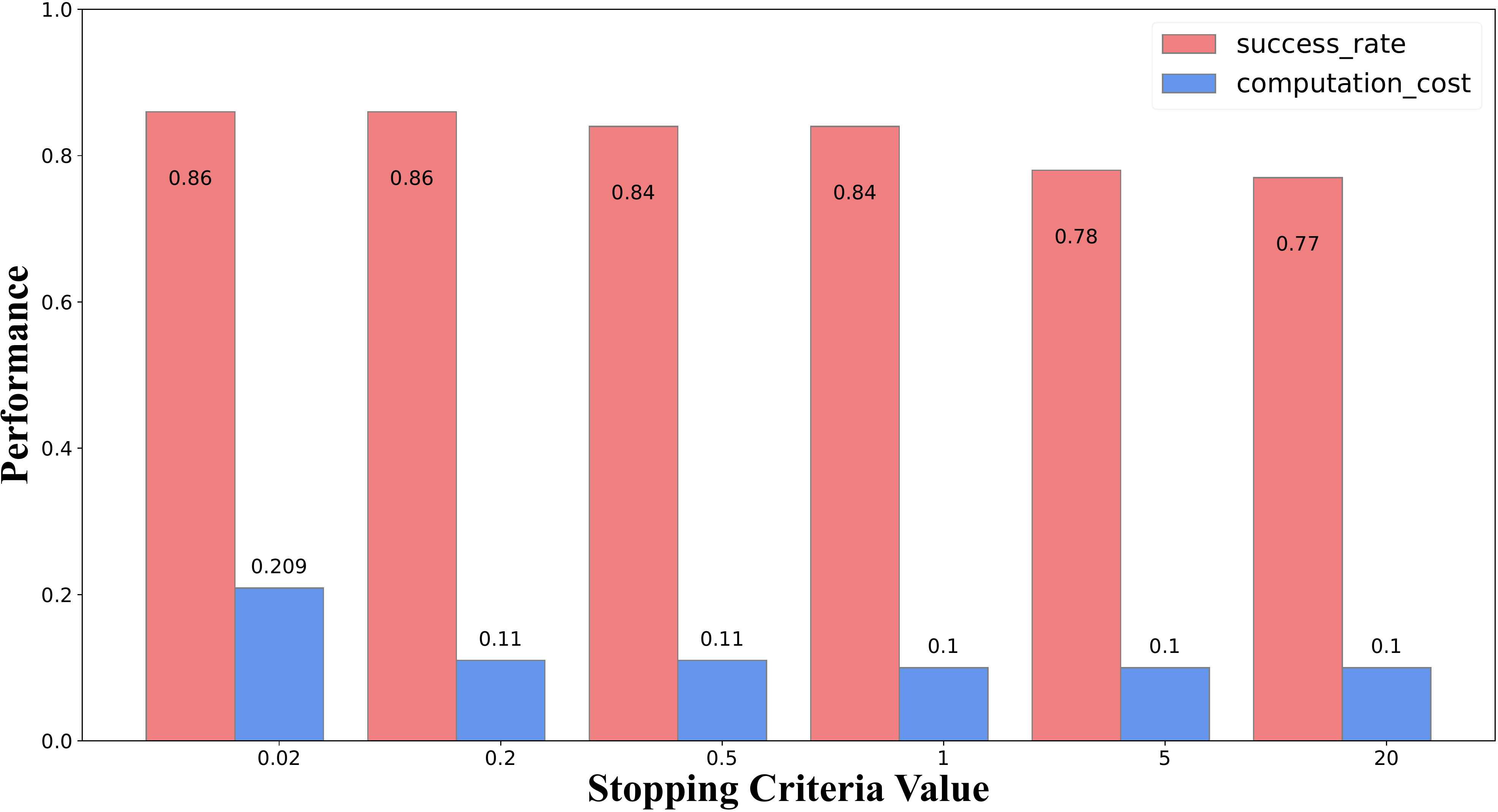}
  \caption{The success rate and computational cost under different stopping criteria values}
  \label{stop}
\end{figure}

To illustrate the relationship between the performance of the algorithm and the stopping criteria, we choose different values of $\epsilon^{\text {pri}}$ and $\epsilon^{\text {dual}}$ for the proposed RDA and execute $50$ trials in the numerical simulator with $8$ randomly distributed obstacles. The average success rate and computational cost are shown in Fig.~10. It can be seen that the success rate and computational cost decrease as $\epsilon^{\text {pri}}$ and $\epsilon^{\text {dual}}$ increase.
But since the success rate drops significantly when $\epsilon^{\text {pri}},\epsilon^{\text {dual}}>1$ and the computational cost 
drops significantly when $\epsilon^{\text {pri}},\epsilon^{\text {dual}}>0.1$, 
the stopping criteria value is chosen to be $\epsilon^{\text {pri}} \in [0.1, 1]$ and $\epsilon^{\text {dual}} \in [0.1, 1]$.

\bibliographystyle{IEEEtran}
\bibliography{reference/Thesis}

\end{document}